\documentclass[letterpaper]{article} 
\usepackage{aaai25}  
\usepackage{times}  
\usepackage{helvet}  
\usepackage{courier}  
\usepackage[hyphens]{url}  
\usepackage{graphicx} 
\usepackage{natbib}  
\usepackage{caption} 
\title{Using Explainable AI and Hierarchical Planning for Outreach with Robots}
\author{
    Rushang Karia\equalcontrib, 
    Jayesh Nagpal\equalcontrib, 
    Daksh Dobhal\equalcontrib, \\
    Pulkit Verma,
    Rashmeet Nayyar,
    Naman Shah,
    Siddharth Srivastava
}
\affiliations{
    School of Computing and Augmented Intelligence\\

    Arizona State University\\
    \{Rushang.Karia,jnagpal1,ddobhal,verma.pulkit,rmnayyar,npshah4,siddharths\}@asu.edu
}

\newcommand{\CC}{\cellcolor{gray!13}}
\usepackage[linesnumbered,ruled,vlined] {algorithm2e}

\SetCommentSty{mycommfont}

\usepackage{subfig}

\usepackage{amsmath}
\usepackage{amsthm}
\usepackage{booktabs}
\usepackage{amsfonts}

\usepackage{multirow}
\usepackage{makecell}

\usepackage[svgnames,table]{xcolor}

\newcommand{\mysssection}[1]{\noindent\textbf{#1}\hspace{5pt}}

\newcommand{\nameAbbr}{JEDAI.Ed}

\usepackage[framemethod=TikZ]{mdframed}
\mdfdefinestyle{PVFrame}{%
    frametitlerule=true,
    frametitlebackgroundcolor=black,
    frametitlefont={\bfseries\color{white}},
    linecolor=black,
    outerlinewidth=0pt,
    roundcorner=0pt,
    innertopmargin=4pt,
    innerbottommargin=4pt,
    innerrightmargin=8pt,
    innerleftmargin=8pt,
    backgroundcolor=black!5}

\usepackage{natbib}

\renewcommand{\cite}{\citep}

\usepackage{pifont}
\newcommand{\cmark}{{\color{Green} \ding{52}}}
\newcommand{\xmark}{{\color{Red} \ding{55}}}

\usepackage{amssymb}
\usepackage{rotating}
\usepackage{tabularx}

\usepackage{pdflscape}

\usepackage{cuted}        

\begin{document}

\maketitle

\begin{abstract}
Understanding how robots plan and execute tasks is crucial in today's world, where they are becoming more prevalent in our daily lives. However, teaching non-experts, such as K-12 students, the complexities of robot planning can be challenging. This work presents an open-source platform, \nameAbbr{}, that simplifies the process using a visual interface that abstracts the details of various planning processes that robots use for performing complex mobile manipulation tasks. Using principles developed in the field of explainable AI, this intuitive platform enables students to use a high-level intuitive instruction set to perform complex tasks, visualize them on an in-built simulator, and to obtain helpful hints and natural language explanations for errors. Finally, \nameAbbr{}, includes an adaptive curriculum generation method that provides students with customized learning ramps. This platform's efficacy was tested through a user study with university students who had little to no computer science background. Our results show that \nameAbbr{} is highly effective in increasing student engagement, teaching robotics programming, and decreasing the time need to solve tasks as compared to baselines.
\end{abstract}

\section{Motivation}
\label{sec:introduction}

\begin{figure*}[!ht]
    \centering
    \includegraphics[width=\textwidth]{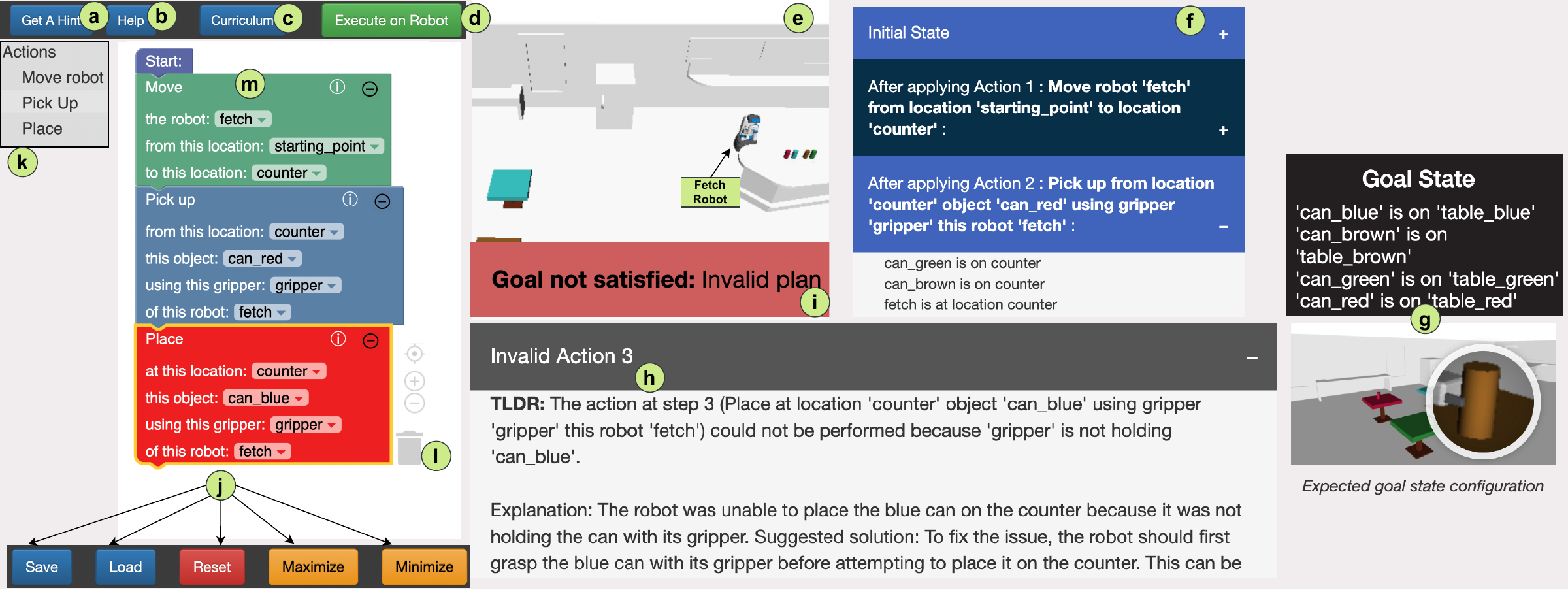}
    \caption{\small A screenshot (best viewed in color) of the \nameAbbr{} user interface (UI) (zoomed in and enhanced for clarity). The annotated circles describe the different sections of the UI (described in Sec.\,\ref{subsec:system_overview}). The supplementary material includes unmodified screenshots along with a video walkthrough of the \nameAbbr{} interface.}
    \label{fig:screenshot}
\end{figure*}

Recent advances in Artificial Intelligence (AI) have enabled the deployment of \emph{programmable} AI robots that can assist humans in a myriad of tasks. However, such advances will have limited utility and scope if users need to have advanced technical knowledge to use them safely and productively. For instance, a mechanical arm robot that can assist humans in assembling different types of components will have limited utility if the operator is unable to understand what it can do, and cannot effectively re-task it to help with new designs.

This paper aims to develop new methods that will allow educators and AI system manufacturers to introduce users to AI systems on the fly, i.e., without requiring advanced degrees in CS/AI as prerequisites. These methods allow for introducing robotics programming to novices. 

\mysssection{Our contribution}  We accomplish our overall objective by introducing \nameAbbr, a web application that abstracts the intricacies of robotics programming and exposes the user to an easy-to-use interface to the robot. \nameAbbr{} incorporates several new features that enable its use in educational settings. Firstly, \nameAbbr{} provides an adaptive curriculum design module that can automatically generate problems catered to a particular user by keeping track of the user's performance. Our system identifies multiple causes of failure and explains them. Finally, \nameAbbr{} utilizes large language models (LLMs) not to discover information but to express factual information and justifications computed using well-defined reasoning processes thereby ensuring the reliability of information being provided.

We implemented \nameAbbr{} by using the existing JEDAI system \citep{shah_2022_jedai} as a baseline. While the core JEDAI system provides a good foundation for development, it has not been developed or evaluated with the components necessary for introductory AI education. E.g., it indirectly requires the users to have some knowledge of robot simulators to operate, does not help educators with designing curricula, etc. Our contributions (mentioned above), along with several other quality-of-life improvements, such as an improved user-interface, etc., make \nameAbbr{} a significant improvement over JEDAI.

We showcase the usefulness of \nameAbbr{} through a user study designed to assess and evaluate its utility and compare it to JEDAI. Our results show that \nameAbbr{} makes robots easy to use and piques curiosity about AI systems. Furthermore, there is a 20\% improvement in solution times and significantly higher positive sentiment compared to JEDAI. Furthermore, we have also piloted \nameAbbr{} in two high-school classes and have received positive feedback showcasing the usefulness of \nameAbbr{} across different age groups.

\section{Background}
\label{sec:background}

In this section, we give a background of key concepts that allow users to program robots for accomplishing tasks.

\mysssection{Running example} Consider a robot that is deployed at a coffee shop to help with its day-to-day operations. Depending upon the day's priorities, the owner may want to program the robot to assist with different tasks such as delivering coffee to customers or washing the cups, etc. To effectively assist the owner, the robot must be able to be ``given" tasks (or instructions) by the owner and autonomously perform them.

\mysssection{Planning} Robots (and humans) often accomplish tasks by computing a fixed sequence of instructions and then executing them sequentially. These sequences are known as \emph{plans}, \emph{planning} is the process of computing such plans, and algorithms that do planning are called \emph{planners}. Planners take an input task and instruction set and output a plan (consisting of instructions from the instruction set) that solves the task. A \emph{valid} plan for a task is a sequence of semantically consistent instructions starting from the initial state.

\mysssection{Robot instructions and motion planning} Robots can only execute a specific type of \emph{low-level} plan known as a \emph{motion plan}. This plan specifies a sequence of movements for each joint of the robot and is obtained using \emph{motion planning}.  E.g., the Fetch robot in Fig.\,\ref{fig:screenshot}e has an arm with $8$ joints: $\theta_1,\ldots,\theta_8$. A motion plan, $\langle [\theta^1_1,\ldots,\theta^1_8], \ldots, [\theta^n_1,\ldots,\theta^n_8] \rangle$, that uses the robot to accomplish an example task of picking up a coffee cup from the counter would contain a sequence of \emph{low-level} instructions $[\theta^i_1,\ldots,\theta^i_k]$ that contain numeric values, $\theta_k \in \mathbb{R}$, for all of its joints. Computing such low-level instructions needs robot-specific knowledge and requires complex algebraic arithmetic to compute a motion plan that provides smooth (and safe) motion. These constraints make motion planning quite difficult for humans.

\mysssection{Human instructions and plans} Contrary to robots, humans typically accomplish tasks by following instructions at a higher level of abstraction than robots.
E.g., to accomplish the same task described in the preceding paragraph, a human often computes a \emph{high-level} plan, $\langle$\texttt{Go to the counter}, \texttt{Pick up the coffee cup}$\rangle$, consisting of \emph{high-level} instructions. Humans can find (and execute) high-level plans for complex tasks fairly easily, however, robots can not use such plans directly to accomplish tasks. 



\mysssection{Hierarchical planning} Given the difficulty of motion planning, it is easy to see that programmable robots must accept high-level instructions to be usable by humans. In this work, we focus on human-in-the-loop (HITL) robot programming where high-level plans are provided by a human and a hierarchical planner converts such plans into a sequence of motion plans that the robot can execute. 

\mysssection{Explaining Failures} This tiered approach to HITL robotics programming introduces some new hurdles. One key challenge is that high-level plans might not be successfully compiled into low-level plans. E.g., a high-level plan 
$\langle$\texttt{Pick up the coffee cup}$\rangle$ cannot be compiled into a low-level plan for a single-arm robot if it is already holding something else. When such failures occur, it is imperative that the robot appropriately informs the user of the failure in high-level terms that the user can easily understand. Explaining why a failure occurred can allow a user to correct (or modify) the high-level instructions so that the desired behavior can be achieved. E.g., an explanation of the form ``\emph{I (the robot) cannot pick up the coffee cup because I am currently holding a water bottle}" allows the user to  (a) identify why the robot could not accomplish the task, and (b) modify their instructions so that the robot can accomplish it.

\section{The \nameAbbr{} Platform}
\label{sec:approach}

We aim to develop a platform that makes robotics programming accessible to a wide spectrum of users and use cases. Thus, we have taken several design considerations (detailed in the supplement) to develop \nameAbbr{}, an open source\footnote{We plan to publicly release the source code post acceptance.} pedagogical tool that brings robotics programming into the hands of novice users. \nameAbbr{} is usable by educators seeking to teach classes on AI, by hobbyists who are interested in robotics, and many others. We compare \nameAbbr{} with JEDAI w.r.t. some of the desiderata in Table\,\ref{tab:jedai_vs_jedai_ed}.

The next section discusses \nameAbbr{}'s features that make it an ideal pedagogical platform for robotics programming followed by an example use case of \nameAbbr{} for programming a robot on a task from our user study.

\subsection{Learning Objectives}
The objective for \nameAbbr{} is to facilitate the understanding of reasoning and quickly provide high-level instructions to robots to perform tasks. Furthermore, our platform explains failures and thus allows users to learn more about the capabilities of the robot. Our focus ties well with the objective of the AI4K12 Big Idea 2 -- Representation \& Reasoning\footnote{\url{https://ai4k12.org/big-idea-2-overview/}} which requires users to be able to reason about how their instructions can change the state of the world and use this knowledge to compute to plan.

\begin{table}[t]
    \centering
    \begin{tabular}{rcc}
         \toprule
         \textbf{Desiderata} & \textbf{\nameAbbr{}} & \textbf{JEDAI} \\
         \midrule
        Open source & \cmark & \cmark \\
        Minimal system requirements & \cmark & \cmark \\
        Integrated simulation & \cmark & \cmark \\         
         Intuitive user interface & \cmark & \xmark \\
         Adaptive problem generation & \cmark & \xmark \\
         Multi-failure explanations & \cmark & \xmark \\
         LLM-powered NL explanations & \cmark & \xmark \\
         \bottomrule 
    \end{tabular}
    \caption{\small A comparison of some of the features of \nameAbbr{} compared to JEDAI. A detailed description of the desiderata is available in the supplementary material.}
    \label{tab:jedai_vs_jedai_ed}
\end{table}

\subsection{System Overview}
\label{subsec:system_overview}

The \nameAbbr{} architecture, illustrated in Fig.\,\ref{fig:architecture}, is modular in design allowing for easy customization (discussed in supplement). Fig.\,\ref{fig:screenshot} shows the overall \nameAbbr{} interface that is presented to users. The \nameAbbr{} user interface module (UI) is the front-end that users interact with and can be run on any modern web-browser making it widely accessible. The back-end can be hosted on any server.

\subsubsection{User Interface (UI)}
The \nameAbbr{} UI follows the single-page application (SPA) design methodology providing the user with
all pertinent information on a single page thereby reducing navigation fatigue. Users are presented with a playground area where they can utilize the intuitive, high-level instruction sets (Fig.\,\ref{fig:screenshot}$k$) to create plans via Blockly \citep{Google_2018_blockly} -- a block-based programming language. For example, the \texttt{Move} action in Fig.\,\ref{fig:screenshot}$m$ represents the high-level instruction `\textit{Move the robot fetch from the starting point to the counter}'. Finally, \nameAbbr{} provides feedback via different modalities (e.g., an audio click when blocks are connected, changing the color of invalid blocks to red, etc.).

\subsubsection{Low-Level Module (MPM)} \nameAbbr{} provides an integrated low-level planner, ATAM \cite{shah_2020_anytime} and simulator, OpenRAVE \citep{diankov10_openrave} for executing user-provided plans on a robot using the UI (Fig.\,\ref{fig:screenshot}$d$). ATAM converts high-level plans to low-level plans that can be executed and visualized on the UI via the simulator (Fig.\,\ref{fig:screenshot}$e$). This execution is a close approximation of the real-world. The motion planning process \ is streamed in real-time providing informative insights about it. We include one such execution in the video walkthrough included in the supplement.

\subsubsection{User Assistance Module (UAM)} It is well-known that iteration and improvement are part of the learning process and learning from failures can be expected in an educational setting \cite{Jackson2022}. \nameAbbr{} uses advances in explainable AI to automatically generate explanations that allow users to (a) learn why their plans are failing, (b) better understand the robot's limitations and capabilities, and (c) fix their plans so that the robot can accomplish the tasks.

\noindent
\textit{Explanations:} Our system uses HELM \cite{sreedharan_2018_helm} and VAL \cite{Howey2004val} for generating explanations whenever failures occur in the user-submitted high-level plan. Once a user connects any block, the current plan is routed through these components to identify whether the plan is valid. An invalid plan is passed to HELM and VAL to generate formal explanations which are then translated to NL via templates and LLMs and displayed to the user.

\begin{figure}[t]
    \centering
    \includegraphics[width=\columnwidth]{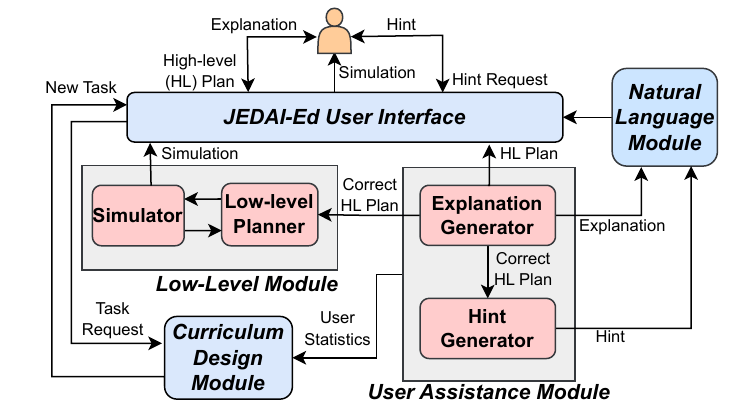}
    \caption{\small The \nameAbbr{} architecture (described in Sec.\,\ref{subsec:system_overview}).}
    \label{fig:architecture}
\end{figure}

\mysssection{Natural language module (NLM)} This module processes messages from all components, converting them to a human-readable message via hand-coded NL templates and/or LLMs. We use well-defined reasoning processes to avoid hallucinations in LLMs by utilizing LLMs primarily as translators and not as reasoners. In our experiments, we used GPT-3.5 Turbo \cite{openai2023gpt3.5}. However, any LLM can be easily configured.  A detailed description of our prompts and NL templates is available in the supplementary material.

\textit{Ethical Considerations and Guardrails:} We limit the potential of LLMs in generating offensive content by (a) using fixed prompts and not free-form chat mode, and (b) using LLMs that are compliant with the OpenAI content policy which dictates the types of responses the LLM can generate. \nameAbbr{} does not include any user-identifiable information in requests to the LLMs nor is such information required for using any feature of \nameAbbr{}.

\subsubsection{Curriculum Design Module (CDM)} 
\begin{figure}
    \centering
    \includegraphics[width=\linewidth]{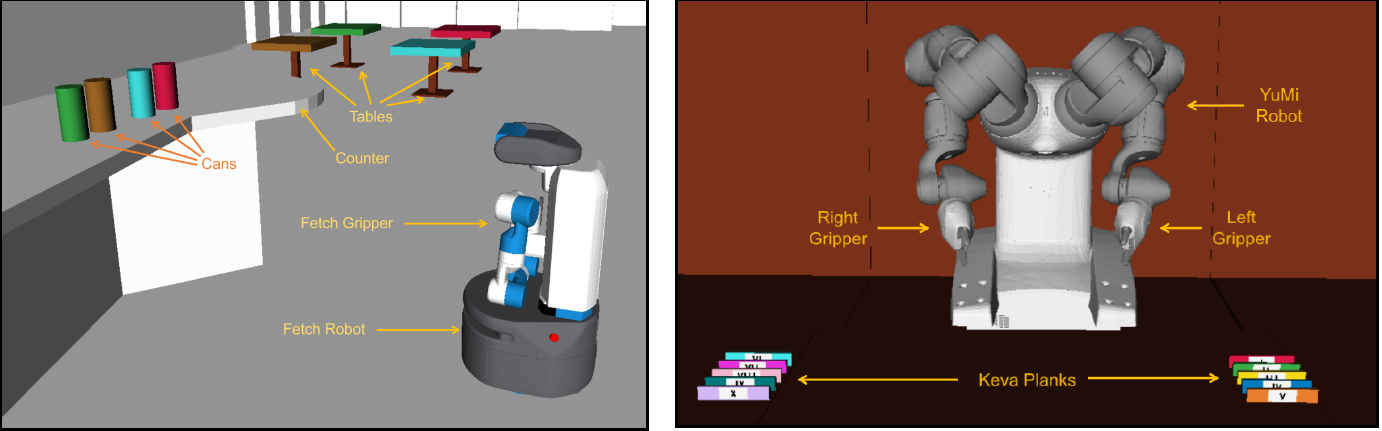}
    \caption{\small Example environments, Coffee Shop (left) and Keva (right), included with \nameAbbr{}. These environments feature sample tasks and problem generators for use as activities. More environments and their details are included in the supplement.}
    \label{fig:packaged_domains}
\end{figure}

\nameAbbr{} includes several environments such as Cafeworld, Towers of Hanoi, etc. that are widely used in AI coursework (Fig.\,\ref{fig:packaged_domains}). These environments provide a diverse mix of tasks and robots that instructors can use as activities for teaching AI planning. Furthermore, \nameAbbr{} also includes a problem generator that can generate new problems on-the-fly by utilizing breadth-first search (BFS) \cite{DBLP:books/aw/RN2020} and allow users to explore the capabilities of the robot on their own.

\textit{Adaptive Problem Generation:} Our platform develops an adaptive problem generation module to help novices understand the capabilities of the robot in a systematic and directed fashion. We do so by keeping track of the user's performance as they solve problems and generating new problems (in the same environment) that focus on aspects that the user has had difficulty with. For example, actions that the user has made mistakes on. The next problem focuses on generating problems that only require the user to use the difficult action thereby reducing the overall cognitive load and making learning easier \citep{teaching_cognitive}. Our overall process for doing so is indicated in Alg.\,\ref{alg:curriculum}. Intuitively, we increase the cost of actions that the user performs well at and decrease the cost of actions that the user has difficulty with. We then use BFS to generate a new problem such that at least one difficult action is covered. The random problem generator described earlier also performs BFS but assumes all actions have equal costs whereas with adaptive BFS, action costs are different and consequently problems generated are not random but directed. This method also works well in a coldstart setting since it initially assumes that the user is not proficient at any action (i.e., $\forall a\text{ }C_u[a] = 0)$. Additional details of this process are included in the supplement.

\mysssection{Walkthrough: Using \nameAbbr{} for a programmable single-arm, mobile robot}
We now describe a typical session of \nameAbbr{} that introduces the functionalities of a mobile manipulator like Fetch (Fig.\,\ref{fig:screenshot}$e$) that is intended to be used in a coffee shop based on the running example (Sec.\,\ref{sec:background}). We also used this in our user-study and the walkthrough describes the typical processes involved. 

First, the educator installs the \nameAbbr{} system on a machine. Next, the educator uses the CDM to select an appropriate environment for the students (e.g., Coffee Shop). The educator then generates (or selects preset) tasks for the student to accomplish (CDM). Alternatively, the educator could instruct the students to use the adaptive problem generator and then solve a test task. The student accesses \nameAbbr{} on a web browser and begins learning. 

The UI presents the user with the necessary information such as the task description and goal (Fig.\,\ref{fig:screenshot}$g$), available instruction set (Fig.\,\ref{fig:screenshot}$k$), and a simulator window (Fig.\,\ref{fig:screenshot}$e$). The goal is provided both in textual as well as visual descriptions along with a magnifier to view finer details of the image. A \emph{Help} button (Fig.\,\ref{fig:screenshot}$b$) provides useful descriptions about the interface and is available to the user at all times. 

The user then uses the instruction sets along with intuitive knowledge to create a plan of high-level instructions by dragging-and-dropping Blockly blocks (Fig.\,\ref{fig:screenshot}$l$) and connecting them to the \emph{Start} block. An audible click lets the user know that the block snapped to another block. 

Every connected block is checked for validity in real-time and explanations (UAM) are provided if the user's current plan contains any invalid actions (Fig.\,\ref{fig:screenshot}$h$). E.g., the explanation shown in Fig.\,\ref{fig:screenshot}$h$ explains that the instruction `\texttt{Place the blue can at the counter using gripper of the Fetch}' failed because the robot was not holding the blue can. The user may also check the result of their current plan in the state display area (Fig.\,\ref{fig:screenshot}$f$). The user may also request a hint (UAM, Fig.\,\ref{fig:screenshot}$a$, elaborated in the supplement) that returns a high-level instruction as a pop-up message. 

Once a valid high-level plan (irrespective of whether it accomplishes the goal or not) is achieved (Fig.\,\ref{fig:screenshot}$g$), the "Execute on Robot'' (Fig.\,\ref{fig:screenshot}$d$) button is activated and the user may submit their plan to be executed on the robot. The planning process and real-time execution of the low-level plan are streamed by the simulator (MPM, Fig.\,\ref{fig:screenshot}$c$).

\begin{algorithm}[!t]
\caption{Adaptive User-Performance Tracking}
\label{alg:curriculum}
\DontPrintSemicolon  

\KwIn{user-performance map $C_u$, action $a$, was-hinted $h$}
\KwOut{Updated user-performance map $C_u$}

$s \gets $ getCurrentState()

\If{canExecuteAction($s, a$) \textbf{and not} $h$} {
    \tcp{User knows action: Cost $\uparrow$}
    $C_u[a] \gets C_u[a] + 1$ 
}
\Else {
    \tcp{User does not know action: Cost $\downarrow$}
    $C_u[a] \gets C_u[a] - 1$
}
\end{algorithm}
\section{Empirical Evaluation}
\label{sec:experiments}

\begin{figure*}
    \centering
    \includegraphics[width=\linewidth]{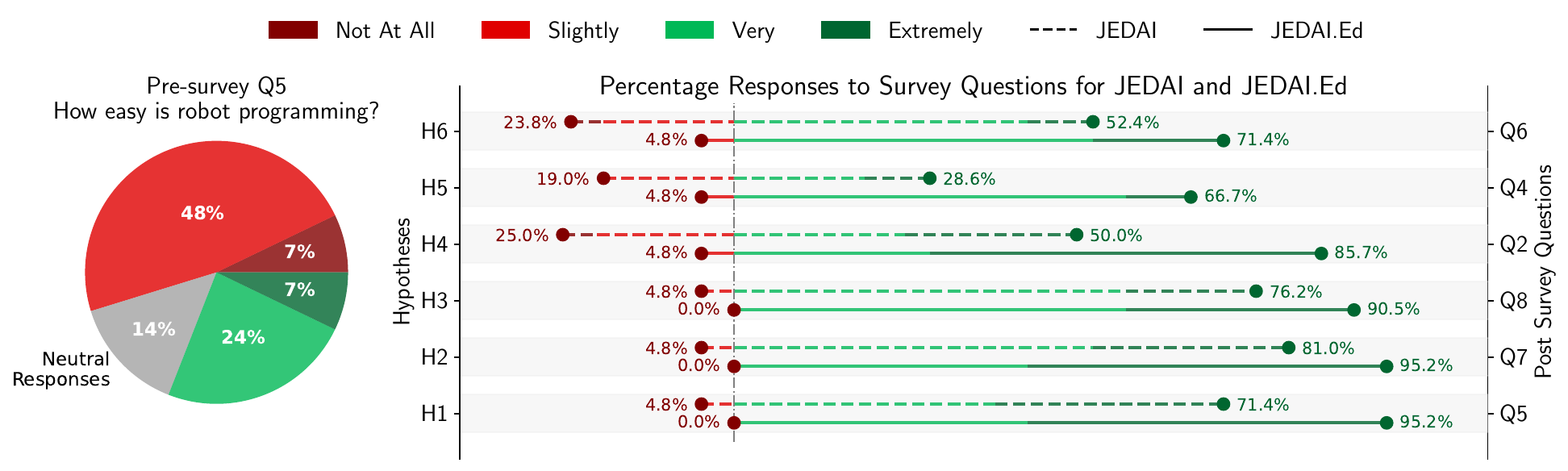}
    \caption{\small Results from our user study with two control groups $(n=42)$ split evenly between \nameAbbr{} and JEDAI. The x-axis plots the responses as a percentage. Green (red) bars to the right (left) indicate positive (negative) sentiment. Stem annotations indicate the total positive (negative) sentiment. The left y-axis specifies the target hypothesis. The right y-axis specifies the question that assesses its validity.}    
    \label{fig:user_study_results}
\end{figure*}

\begin{table*}[h]
    \begin{tabular}[width=\linewidth]{@{}p{0.18\textwidth}@{}p{0.5\textwidth}@{}p{0.01\textwidth}@{}>{\centering}p{0.1102\textwidth}@{}>{\centering}p{0.00\textwidth}@{}>{\centering}p{0.00\textwidth}c@{}m{0.00\textwidth}@{}>{\centering}p{0.011\textwidth}}
    \toprule
    \textbf{} & \textbf{} & & \multicolumn{4}{c}{\textbf{$\mu \pm \sigma$ w.r.t. 1-sample t-test ($\mu^1_0 = 2$)}} &\\ 
    \cmidrule{4-7}
    \textbf{Hypothesis} & \textbf{Post Survey Question used for Testing Hypothesis} & & \textbf{\nameAbbr{}} & & & \textbf{JEDAI} & \\ 
    \midrule
        \makecell[lt]{\textbf{H1:} Increased \\ curiosity} & \textbf{Q5:} As compared to before participating, how much has your curiosity increased to learn more about AI systems and robots? & & \multirow{2}{*}{3.47$\pm$ 0.60} &  &  & \multirow{2}{*}{3.00 $\pm$ 0.89} &\\
    \midrule
			 \makecell[lt]{\textbf{H2:} Easier \\ programming} & \textbf{Q7:} Do you agree that the \nameAbbr{} system made it easier for you to provide instructions to a robot for performing tasks? & &  \multirow{2}{*}{3.47 $\pm$ 0.60} &  &  &  \multirow{2}{*}{3.04 $\pm$ 0.80} &\\
    \midrule
          \makecell[lt]{\textbf{H3:} Improved \\ understanding} & \textbf{Q8:} Do you agree that \nameAbbr{} helps improve the understanding of the robot’s limitations and capabilities? & &  \multirow{2}{*}{3.23 $\pm$ 0.62} &  &  &  \multirow{2}{*}{2.90 $\pm$ 0.76} &\\
    \midrule
          \makecell[lt]{\textbf{H4:} Helpful \\explanations} & \textbf{Q2:} How helpful were the explanations that were given for the cause of an error? & & \multirow{2}{*}{3.38 $\pm$ 0.86} &  &  & \multirow{2}{*}{{\color{red} \textbf{2.42 $\pm$ 1.20}}} &\\
                \midrule
			 \textbf{H5:} Intuitive UI & \textbf{Q4:} How intuitive was the interface? & & 2.71 $\pm$ 0.71 &  &  & {\color{red} \textbf{2.19 $\pm$ 0.87}} &\\
    \midrule
         \makecell[lt]{\textbf{H6:} Programming \\confidence} & \textbf{Q6:} How well do you think you now understand how one can use an AI system to make a plan for a robot to perform a task? & & \multirow{2}{*}{2.85 $\pm$ 0.79} &  &  & \multirow{2}{*}{{\color{red} \textbf{2.33 $\pm$ 1.06}}} &\\
    \bottomrule
    \end{tabular}
    
 \caption{\small \nameAbbr{} user study results $(n=42)$ used to validate our hypotheses. The table provides a short description of the target hypothesis, the corresponding questions used to validate it, and the one-sample t-test results. All results are statistically significant ($p < 0.05$) except for entries in {\color{red} \textbf{bold; red}}. Comprehensive statistical data is available in the supplement.}
 \label{fig:survey}
\end{table*}

We developed \nameAbbr{} to expose novice users to AI and robotics. We conducted a user study to evaluate if \nameAbbr{} achieves the goal by evaluating the following hypotheses:

\mysssection{H1 (Increased curiosity):} \nameAbbr{} increases the curiosity of users to learn more about robotics and AI.

\mysssection{H2 (Easier programming):} \nameAbbr{} makes it easy for users to provide instructions to robots.

\mysssection{H3 (Improved understanding):} \nameAbbr{} improves user understanding w.r.t. the limitations/capabilities of a robot.

\mysssection{H4 (Helpful explanations):} \nameAbbr{}'s provided explanations help users understand (and fix) errors in their plans.

\mysssection{H5 (Intuitive UI):} \nameAbbr{}'s UI is intuitive and easy to use requiring little to no study facilitator intervention.

\mysssection{H6 (Programming confidence):} \nameAbbr{} increases users confidence in instructing robots to accomplish tasks.

\mysssection{H7 (Faster solving):} \nameAbbr{} allows users to solve tasks faster than JEDAI.

To evaluate the validity of these hypotheses, we designed a user study for evaluating \nameAbbr{} and comparing it with JEDAI. We present the study methodology below.
    
\subsection{User Study Setup}
We hired 43 university students with no background in computer science as participants for an IRB-approved user study. We discarded 1 incomplete/invalid response, resulting in a sample size of 42. 23 of these were from a non-STEM background. We divided the participants into two control groups. The first (second) control group was assigned the \nameAbbr{} (JEDAI) system for use in the study. The study lasted 45 minutes, was conducted in-person, and had four phases:
\\
\textit{Pre-survey phase (8 min):} Participants were presented with an introductory video about AI. Next, to acquire a detailed understanding of the participant's background, interests in AI, level of awareness and engagement with AI technologies, we employed a pre-survey questionnaire.
\\
\textit{Training phase (12 min):} This phase was intended to get users familiarized with the system and tasks. Communication with the study facilitator was allowed. Participants were presented sequentially with three tasks of the \emph{Coffee shop} environment (Sec.\,\ref{sec:background}) each of which involved utilizing a Fetch robot to deliver cans to tables. \nameAbbr{}  used the adaptive problem generation algorithm to generate training tasks. We used randomly generated training tasks for JEDAI. We ensured that all generated training tasks needed 50\% fewer instructions to accomplish than the test task.
\\
\textit{Test phase (12 min):} The participants solved a test task during this phase. The test task was much harder than the training tasks and required users to deliver multiple cans (optimally using 16 high-level instructions). No communication with the study facilitator was allowed during this phase. The participants were then asked to complete a post-survey questionnaire whose questions were designed to obtain the participant's opinion on the platform they interacted with and to determine if their interest and curiosity had increased post-use. We also collected system logs for analytics data.
\\
\textit{Sentiment change phase (13 min):} This phase is intended to analyze the sentiment change after interacting with both \nameAbbr{} and JEDAI. In this phase, participants who interacted with \nameAbbr{} (JEDAI) in the previous phases were asked to interact freely with JEDAI (\nameAbbr{}). They were once again asked to answer a post-survey questionnaire. This questionnaire was the same as that of the test phase but they could not see the previous responses. 

\mysssection{Questionnaire methodology}
All responses to the questions used the Likert Scale~\cite{likert1932technique} to provide a more intricate depiction compared to binary responses.

\mysssection{Hypothesis testing} Our Likert Scale data was converted to values from 0 to 4 with 0 (4) being the most negative (positive) response and 2 being neutral. We used the p-value obtained by using the one-sample t-test \cite{Ross2017onesamplettest} to test the statistical significance. Within a control group, we assumed the data to be two-tailed for the questions used to validate the hypotheses and used the hypothesis of no difference, i.e., $\mu^{1}_0$ = 2 (balanced Likert scales), as the null hypothesis. Thus, for statistically significant results $\mu^{1}_0 > 2$ denotes positive sentiment and vice versa. 


\subsection{Study Results}

\begin{figure}[t]
    \centering
    \includegraphics[width=\columnwidth]{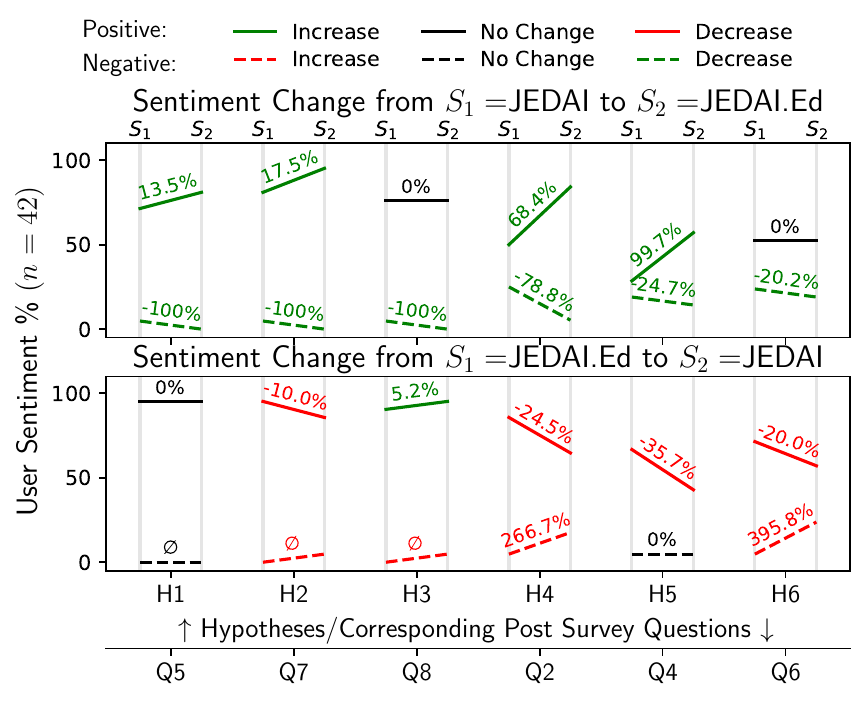}
    \caption{\small Slope charts for results from our sentiment change phase. Users interacted with $S_1$ first and $S_2$ next. The y-axis shows the absolute user sentiment while annotations on the line plots show the \% improvement $(\frac{S_2 - S_1}{S_1} \times 100)$. We use $\varnothing$ when $S_1=0$.}
    \label{fig:sentiment_change}
\end{figure}

Fig.\,\ref{fig:user_study_results} and Table.\,\ref{fig:survey} show our results, the survey questions used to analyze the hypotheses, and data from the statistical tests. All data was statistically significant for \nameAbbr{}. Moreover, \nameAbbr{}'s $\mu \ge 2.7$ showcases an improved, positive experience. We analyze our results below.

\mysssection{H1 (Increasing curiosity):} Fig.\,\ref{fig:user_study_results} shows that after interacting with \nameAbbr{}, user curiosity is 95\% positive. This is far greater than JEDAI, whose positive user sentiment is 71\%. 

\mysssection{H2, H3, and H6:} Our pre-survey results (Fig.\,\ref{fig:user_study_results}, pie) show that before using \nameAbbr{}, 55\% of users believed that robot programming was not easy.
\\
\textit{H2 (Easier Programming):} 95\% of users thought that \nameAbbr{} made it easier to program robots. In contrast, JEDAI only managed to increase positive sentiment to 71\%.
\\
\textit{H3 (Improved understanding), H6 (Programming confidence):} After interacting with \nameAbbr{}, 90\% of users think that they better understand the robot's capabilities, and 71\% of users were confident that they could program robots. 

\mysssection{H4 (Helpful explanations):} Users were extremely positive in their feedback w.r.t. \nameAbbr{} provided explanations ($\approx$86\% of users thought that the explanations were helpful). As compared to JEDAI, \nameAbbr{} provides both brief and LLM-based descriptive explanations that better explain why a failure occured. Thus, JEDAI explanations were rated significantly lower and also had a 25\% negative sentiment.

\mysssection{H5 (Intuitive interface):} Most users using \nameAbbr{} were able to navigate the interface without any help. \nameAbbr{} is modern and includes many quality-of-life features such as the ability to minimize blocks, etc. which are lacking in JEDAI. Fig.\,\ref{fig:user_study_results} shows that 66\% of users found \nameAbbr{}'s UI intuitive as compared to JEDAI which had only 28\% positive sentiment and had a 19\% negative sentiment.

\begin{figure}[t]
    \centering
    \includegraphics[width=\columnwidth]{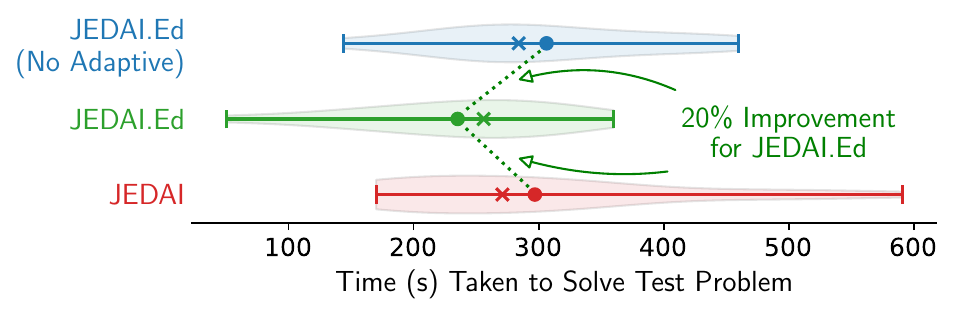}
    \caption{\small Violin plots that indicate the time needed to solve the test task. {\color[HTML]{2ca02c} $\bullet$} ({\color[HTML]{2ca02c} $\times$}) represents the mean (median).}
    \label{fig:solve_times}
\end{figure}

\mysssection{H7 (Faster solving):} Fig.\,\ref{fig:solve_times} shows the distribution of times required to solve the tesk task. \nameAbbr{} users were able to solve the test task in 235 seconds which is 20\% faster than JEDAI. There were 4 (3) users for \nameAbbr{} (JEDAI) that were not able to solve the test task. One additional \nameAbbr{} user encountered an internal error requiring a system restart thus resulting in them not being counted.

One of the key advantages of \nameAbbr{} is the adaptive problem generation that appropriately adjusts the difficulty of tasks so that users can learn faster. \nameAbbr{} also informs users of multiple invalid actions (and explaining two of them) in real-time as compared to JEDAI where users need to submit plans to get any feedback and are only informed and explained of a single invalid action. 

\textit{Ablation Study} To further investigate the impact of adaptive problem generation, we conducted an ablation study by recruiting an additional 19 students with similar backgrounds. These students were administered the same study using \nameAbbr{}. The only change we made was to use randomly generated problems in the training phase instead of the adaptive problem-generation method employed earlier. Three users in this new study were unable to solve the test task. Our results in Fig.\,\ref{fig:solve_times} show that without the adaptive problem generation, the training tasks are much harder for the students and consequently they cannot perform as well on the test task. We attribute the similarities between the solve times w.r.t. JEDAI to the fact that JEDAI also explains failures and thus provides similar feedback.

\subsubsection{Improved Sentiment over JEDAI} Fig.\,\ref{fig:sentiment_change} shows that users have a positive (negative) sentiment change across all metrics when interacting with JEDAI (\nameAbbr{}) first and then experiencing \nameAbbr{} (JEDAI). These observations, along with the rest of our analysis, shows that \nameAbbr{} offers several significant improvements over JEDAI resulting in an overall enhanced user-experience when using \nameAbbr{} as a platform for robotics programming.
    
\subsection{Pilot Program on High School Students}
\begin{figure}[t]
    \centering
    \includegraphics[width=\linewidth]{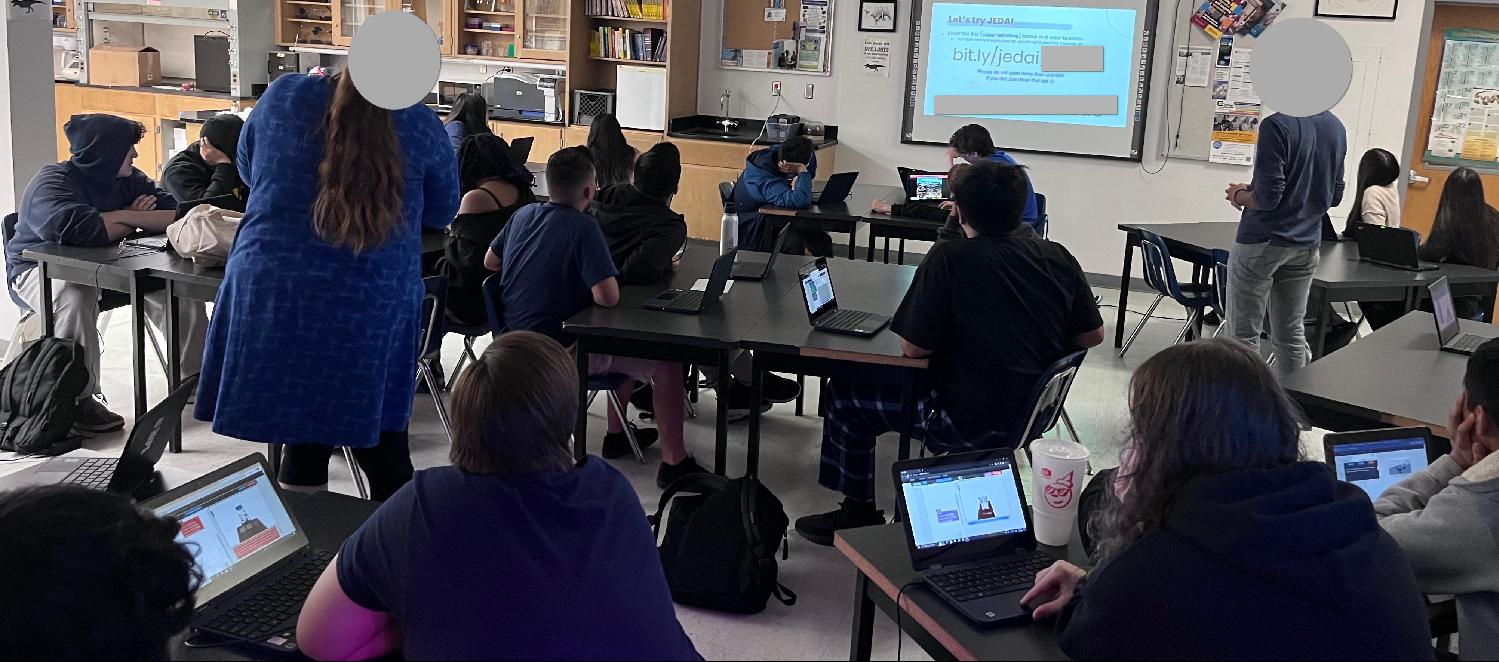} \\
    \includegraphics[width=\linewidth]{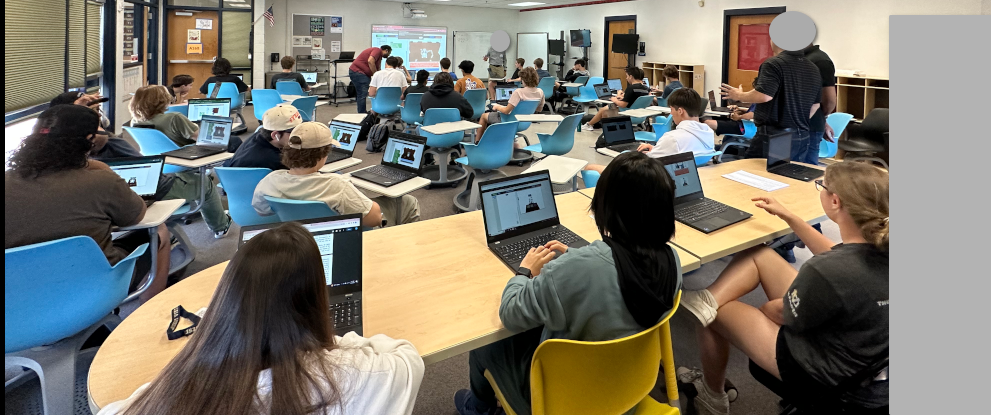} \\
    \caption{\small Engagement from \nameAbbr{}'s pilot program on two high schools.}
    \label{fig:pilot_program}
\end{figure}

We also demonstrated \nameAbbr{} across 3 sessions at two different high schools to a total of $\approx$90 students. Each session lasted 60 minutes and students were asked to complete tasks across 3 different environments (Coffee shop, Keva $\pi$--Planks, and Towers of Hanoi). Most students were able to complete all tasks without any supervision.
Pictures from our visits are included in Fig.\,\ref{fig:pilot_program}. We also solicited feedback from the program coordinator who mentioned:

\textit{``They found it very user-friendly. Thank you again for the visit and looking forward to seeing more in the future."}

\subsection{Improvement Opportunities} 
We now discuss what didn't work, and improvement opportunities based on feedback from the user study and our pilot. 

For our pilot program, we hosted our system on an AWS cloud instance to serve the students. The network latency between the school and the server was visible in the interface and caused latency issues where the video stream of the robot executing the plan was not rendering correctly. Optimizing the motion planner to break down the trajectory packets and send them piece-by-piece would provide for a smoother experience which we are currently implementing.

Some users from had difficulty understanding that plans begin at the \textit{Start} block. They mentioned that renaming it to \textit{Connect blocks here} would improve the UI's intuitiveness.

We observed that in its current iteration \nameAbbr{} is not widely accessible on devices which do not employ a keyboard and mouse. We plan to improve the accessibility of our platform by using generative AI so that users can provide plans verbally using multimodal models.

An additional feature that we are working on allows users to use programming constructs like loops and conditionals to form their plans. These allow for the inclusion of richer environments with non-deterministic action semantics. Explaining failures in such programs is a challenging and exciting question for future research. 

\section{Related Work}
\label{sec:related_work}

This work brings together several independent research directions in a single platform. We discuss them here.

\mysssection{Visualizations in planning}
There are tools that help visualize the planning process to make it is easy to understand for the users. Such tools include Web Planner~\cite{magnaguagno_2017_web}, Planimation~\cite{chen2020planimation}, PDSim~\cite{dePellegrin_2021_pdsim}, vPlanSim~\cite{Roberts_2021_vplansim}, PlanVis \citep{cantareira2022actor} etc. 
These methods focus on visualizing the planning process for users, whereas \nameAbbr{}, in addition, also helps novices in planning on their own, executing the plans on robots, and explaining their mistakes to them.

\mysssection{Robot programming interfaces} 
CoBlox~\cite{Weintrop_2018_evaluating} used a similar interface for creating low-level plans for robots but also requires users to provide low-level plans. \citet{winterer_2020_expert} analyzed the use of Blockly for programming industrial robots. 
These approaches target expert users and unlike \nameAbbr{} cannot be used by novices.

\mysssection{AI concepts for students} 
Robot-VPE~\cite{Krishnamoorthy_2016_using} and Code3~\cite{huang2017code3} used a Blockly-like interface for K12 students to write programs for robots. \citet{broll2023beyond} created a tool to teach complex ML concepts to students using block-based pre-programmed games. Maestro~\cite{eleta2023maestro} used goal-based scenarios to teach students about robust AI.

\mysssection{Generating explanations with easy-to-understand interfaces} There is a large body of work on generating explanations for user-provided plans. Few such approaches~\cite{grover_2020_radar,Karthik_2021_radarx,brandao2021towards,kumar2021vizxp} use an easy-to-understand user interface and natural language to make the explanations easily accessible to novice users. These approaches do not integrate low-level planning and thus cannot be used to program robots.

\section{Conclusion}
\label{sec:conclusion}

\paragraph{} 
We introduced \nameAbbr{}, an open-source platform to introduce high-level robot planning to novices.
We showed that \nameAbbr{} is an effective and intuitive platform in teaching AI planning to users without a background in the subject. \nameAbbr{} significantly improves upon its predecessor and adds several new and novel features. Adapting curriculums tailored to individual users allows for more effective learning which is evident from faster solution times on our platform. Our results show that users prefer \nameAbbr{} over JEDAI.  Moreover, \nameAbbr{} was able to successfully engage students and pique their curiosity in learning more about AI planning. Our pilot program was highly successful and increased the students' confidence in robotics programming. We hope to keep developing and making \nameAbbr{} available to wider audiences.

\section*{Acknowledgements}
This work was supported in part by the ONR under grant N000142312416.
\section*{Ethical Statement}
This work involved recruiting humans for our study. Both our pre and post survey questionnaires went through an IRB review process and were approved before starting the study. We ensured that students had little to no background in computer science and only allowed participants who either (a) were not enrolled in a computer science major, (b) did not have any significant programming experience, and (c) had not formally or informally enrolled in a data structures or equivalent course either through a university or an online education platform. For computer science majors, data structures is a pre-requisite for a majority of programming and robotics related classes. Thus, our computer science majors were composed mainly of students in their freshmen year with little to no exposure to any computer science concepts.

Usage of LLMs carries the risk of providing content that might not be relevant or might be offensive to its users. We mitigated this by using OpenAI's latest GPT-3.5-turbo model (\texttt{gpt-3.5-turbo-0125}) which is compliant with the OpenAI usage policy \cite{openaiusage} on content generation. LLMs are more prone to generate irrelevant or offensive content when engaged in a dialogue with users. In our case, our prompts are structured and fixed and thus are unlikely to generate irrelevant text. Additionally, in accordance with the company policy, GPT-3.5 has default content filters that stop any offensive or inappropriate text from being generated and returned to be displayed in \nameAbbr{}.

Finally, with regards to user privacy, no user identifying information was provided to GPT-3.5 or used at any point in \nameAbbr{} and in our experiments with JEDAI.

\bibliography{aaai25}

\clearpage
\appendix

The supplementary material begins with an in-depth treatment of the desiderata mentioned in the main paper. Also included, is a video walkthrough of some of the features of the interface. This is followed by a exposition of the design considerations and extensibility of \nameAbbr{}. We then briefly introduce environment domains that are included in \nameAbbr{} and conclude with details about the user study and analysis of the results. The large size of the source code prohibits us from including it in the supplementary material, but we plan on releasing \nameAbbr{} as an open source software in case of acceptance (as we mentioned in Sec.\,\ref{sec:approach}).

\section{Desiderata}
\label{sec:desiderata}
Table \ref{tab:desiderata_supplement} provides a contrast between the features of \nameAbbr{} and JEDAI.

\subsection{Intuitive user interface} \label{ssec:intuitive_ui} The JEDAI user interface was a static browser window which would display all the user assistance information on to the screen simultaneously. The result was a text dense webpage. Furthermore, JEDAI required the user to manually click a button in order to check the validity of their plans and receive explanations about their errors. To test even incremental changes in their plans, users would have to move the mouse pointer back and forth between the Blockly workspace and the submit button. Finally, JEDAI presents the robot simulation in a separate browser window, further burdening the user with managing multiple browser tabs or windows while solving tasks.
\paragraph{} The unmodified screenshots for \nameAbbr{} and JEDAI are presented in Fig.~\ref{fig:unmod_jedai_ed} and Fig.~\ref{fig:unmod_jedai} respectively. 

\subsection{UI Improvements} \label{ssec:ui_improvements}Extensive work has gone into remodelling the interface of JEDAI to be more user-friendly and intuitive. An important motivation behind this was to not overload the user with too much information on the screen and bring all the components, including the simulation window into a single browser window. Optimizing on the screen space available, \nameAbbr{} presents explanations, hints etc. as collapsible text fields. This allows for a high level summary to be visible immediately on the screen along with the option for the user to expand and consume more detailed information if needed. In order to make the connection between an error explanation and the corresponding block clear to the user, \nameAbbr{} also highlights the action block when its corresponding explanation dropdown is selected. Similarly the robot simulation was integrated within the same webpage containing the Blockly workspace and explanation dropdown lists, thus providing a single, compact view port for the user. 

\subsection{Blockly} \label{ssec:blockly_ui}Another accessibility feature introduced in \nameAbbr{} is the ability to minimize and maximize blocks. Tasks that require longer plans, the connected blocks quickly overflow beyond the workspace requiring the user to scroll. Minimizing previously added blocks allows the user to be able to add and edit the arguments of the new block they are adding in the same workspace without needing to scroll down to find the last block each time.

\subsection{State Space Display} \label{ssec:ss_display}To facilitate the users' understanding on the environments that they are solving tasks in, \nameAbbr{} introduces a State Space Display. Starting with the initial state of the environment and listing the change after the application of each action n the users submitted plan, the list of all true predicates is displayed, as seen in Fig.~\ref{fig:screenshot}$f$. Again, in keeping with the objective of not filling up the screen with text dense messages, the list of predicates is hidden by default and the user can expand and collapse the list as they need. In the event of an invalid submitted plan, the display is truncated at the last valid action and the error explanation dropdown is displayed.

\subsection{Curriculum generation} \label{ssec:curriculum_generation} 
\paragraph{}

JEDAI and \nameAbbr{} both contain a diverse set of domains and problem tasks for users to solve (App.\,\ref{sec:bundled_environments}), however JEDAI does not offer an in-built method which provides a systematic learning journey to the user. In JEDAI, an educator would have to manually design a lesson plan with students sequentially exposed to problems of increasing difficulty, but that would still be a one-size-fits-all approach, without possibility of the curriculum being tailor-made for individual learners' needs. This problem is alleviated by \nameAbbr{}'s curriculum design module which automaticallyadapts to each individual learner and provides them a customized learning path for learning about the capabilities of a robot in an environment. Our curriculum module progressively generates challenging problems of increasing difficulty, while automatically adapting to users performance.

\paragraph{}User created plans are continuously evaluated to assess their level of understanding of the various actions. Each plan is broken down into its constituent actions which are sequentially applied starting from the initial state. An action being applicable in the current state implies the user having a correct understanding of said action. Here, we increment the cost associated with that action and update the current state to be the state resulting in the application of the action on the current state. Conversely, if an action is inapplicable, the entire plan is rendered invalid and updating the current state is impossible. The associated cost of the action is decremented and the action-cost mapping is returned. This above process is illustrated in Alg. 1.




\paragraph{} Alg.2 showcases our overall process for adaptive problem generation. To generate a problem task using this action-cost mapping, we use a simple greedy search over the state space, where nodes are states of the environment and weighted edges are actions with the weight of the edge being the cost associated with that action. The fringe is initialized as a minimum priority queue, with the priority of a node set as the cumulative action cost needed to reach that node and the order of addition to the fringe used to break ties. Nodes are then popped from the fringe and for each popped node, the states reachable from that node in one action are added to the fringe with the incremented cumulative action-cost. This procedure is carried out until an action currently unknown to the user is seen, or if a preset maximum depth is reached. For our experiments, the maximum depth was set to four.

\begin{algorithm}[ht]
\caption{Adaptive Curriculum Task Generator}
\DontPrintSemicolon  

\KwIn{adaptive action-cost mapping $A$, initial state $s_0$, set of grounded actions $\mathcal{A}$, maximum tree depth $d_\textit{max}$}
\KwOut{Problem Goal State $g$}

$f \gets \textit{Min-Priority-Queue}$\;
$v \gets \textit{Empty visited set}$\; 
$\textit{add}(f, (s_0,0,0))$\;
$\textit{add}(v, s_0)$\;

\While{True}{
    $s, c, d \gets \textit{pop(f)}$\;

    \ForEach{$a \in \mathcal{A}$} {
        
        \If{$\textit{isApplicable}(s, a)$} {
            $s' \gets applyAction(s,a)$\;
            $c' \gets \textit{c + A[a]}$\;
            $d' \gets \textit{d + 1}$\;
            \If{$A[a]=0 \textit{ or } d'\ge d_\textit{max}$} {
                $g \gets \textit{s}$\;
                \KwRet{$g$}\;                
            }
            \If{$s' \not\in  \textit{v}$} {
                $\textit{add(v,}s')$\;
                $\textit{add(f, (s',d',c'))}$
            }            
        }
    }

}
\label{algo:currculum-gem}
\end{algorithm}

\paragraph{} All action costs are initialized uniformly to zero, representing that the user is unfamiliar with all the actions. As the user solves curriculum generated problems correctly, they are presented with problems that each introduce an additional unknown action. This ensures that while solving a problem, the user has only one new action to learn thus easing the cognitive burden placed upon them. The very first curriculum task generated, with all action costs set to zero, will select any action applicable in the initial state and return the state resulting from its application as the goal state. Future work may investigate how this cold start problem impacts user engagement and learning, and how it may be resolved.

\subsection{Intuitive Explanations and Hints} \label{ssec:intuitive_expl_hints}In explaining action failures, JEDAI relied upon hand coded text templates, which print out the failing action and the unmet preconditions of said action. While logically sound and providing complete information, theses explanations are not very human friendly to read. \nameAbbr{} uses GPT-3.5 to convert these explanations to be more human readable and user friendly. JEDAI is also limited to explaining failure for only one action, which is a handicap \nameAbbr{} does not suffer from.
\paragraph{} \nameAbbr{} also allows users to get hints to help them in high-level planning. The hint is displayed in the form of a grounded high-level action, with the action name visible, but some of the grounded objects that are arguments to the action obscured. This achieves the twin objectives of nudging the user in the right direction while also not spoon feeding them the answer directly.

\textit{Hint Generation:} \nameAbbr{} comes equipped with FF \cite{Hoffmann_2001}, a \emph{fast} high-level planner. FF can be used to generate the next high-level instruction needed to accomplish the task. This is presented as a user-interpretable hint to the user. Since high-level planning is a hard problem \cite{DBLP:journals/ai/Bylander94}, hinting comes preconfigured with a timeout and displays a status message if hints cannot be computed within the time limit.

\subsection{LLM Prompts} \label{ssec:llm_prompts} We use LLMs to translate explanations and hints to be more readable and user friendly. The explanation or hint generated from the user assistance module is augmented with domain and problem pddl files of the problem being solved. This information, along with a simple prompt is fed into the LLM which is tasked with a translating the explanation given the context to a readable natural language description. The LLM itself does not generate any explanation or hint, but merely acts as a translation interface. 
\paragraph{} GPT-3.5 was prompted with the following text to generate user-friendly explanations and hints, which is domain agnostic and through repeated experimentation, was found to generalize across multiple problems and domains. Text within angular braces acts as a placeholder for the actual text that is input to the prompt, but omitted here for brevity. Note that the LLM is not being tasked with generating any explanation, but only to convert them to natural language: 
\begin{mdframed}[style=PVFrame,frametitle={Explanation Generation Prompt}]
    The following lines describe the $\langle$domain$\rangle$ domain file :\\
    $\langle$domain\ pddl$\rangle$\\
    \\
    The problem to be solved is described in pddl format  as:\\
      $\langle$problem\ pddl$\rangle$\\
    \\
    While running a plan for a problem, an action failed and an explanation generator was used to generate the following explanation:\\
    Explanation:  $\langle$explanation$\rangle$\\
    \\
    The state of the problem - which means the set of predicates that are true in the plan upto the first invalid action are as follows \\
    State: $\langle$state$\rangle$\\
    \\
    Can you please convert the explanation into a brief, more non-expert friendly message that a novice user can understand?
    Also, can you suggested briefly what could be done to fix the issue, taking into account the state reached by the plan so far? 
\end{mdframed}

\paragraph{}The error explanation for a failing action, generated using fixed text templates, as seen in  Fig.\,\ref{fig:screenshot}$h$ (marked as the TLDR version) is presented below:

\begin{mdframed}[style=PVFrame,frametitle={Text Template Based Explanation}]
The action at step 3 (Place at location 'counter' object 'can\_blue' using gripper 'gripper' this robot 'fetch') could not be performed because 'gripper' is not holding 'can\_blue'.
\end{mdframed}

\paragraph{} Using the explanation generation prompt we mentioned earlier, and filling in the above text template generated explanation, GPT-3.5 generated the following user friendly explanation, which is also depicted in Fig.\,\ref{fig:screenshot}$h$ (marked Explanation) :

\begin{mdframed}[style=PVFrame,frametitle={LLM Generated Explanation}]
Explanation: The action of placing the blue can on the counter failed because the gripper is not holding the blue can. To fix this issue, you can have the robot fetch the blue can and place it on the counter using the gripper properly before trying to place it on the table. Make sure the gripper is holding the correct object before attempting to place it at the desired location.
\end{mdframed}


\subsection{Prompt Engineering} \label{ssec:prompt_engineering}Educators can customize LLM descriptions for audiences of different expertise by simply designing appropriate prompts and using them as input to \nameAbbr{}'s fine-tuned LLM without any additional training. This allows educators to take into account the needs of different users without having to expend resources for training or fine-tuning an LLM. An example prompt that can be used for generating explanations for novice users is stated below.
\begin{mdframed}[style=PVFrame,frametitle={Hint Generation Prompt}]
    This is the pddl domain file for the $\langle$domain$\rangle$ domain :\\
    $\langle$domain pddl$\rangle$\\
    \\
    A user has to solve a this problem task described in pddl \\
    $\langle$problem pddl$\rangle$\\
    \\
    The plan was run till the problem reached this state - that is the set of predicates that are true :\\
    $\langle$state$\rangle$\\
    \\
    And the hint generated, which suggests which next action to take with certain arguments to actions replaced with ? is given below:\\
    $\langle$hint$\rangle$\\
    \\
    Can you please convert the explanation into a brief, more non-expert friendly message that a novice user can understand?
    Also, can you suggested briefly what could be done to fix the issue, taking into account the state reached by the plan so far?
\end{mdframed}

A sample hint generated using hand-coded text templates is illustrated below: 

\begin{mdframed}[style=PVFrame,frametitle={Sample Hint}]
You might want to try the action: Move To Counter from this location ? the robot ? to this location counter
\end{mdframed}


The LLM generated counterpart is shown below. In a fashion similar to the explanation translation, the LLM here is used to translate and paraphrase the  hints and solutions using the hint, domain and problem pddl and the state of the environment passed as context:

\begin{mdframed}[style=PVFrame,frametitle={LLM Generated Hint}]
The problem is to solve a task in the Coffee Shop domain. The current state of the problem is that the gripper is empty, there are orders for different cans at the counter location, and the robot is at the starting point. The goal is to place the red can on the red table. To solve the problem, the next action that should be taken is to move the robot to the counter location. This action requires specifying the current location of the robot and the counter location. To fix the issue, you need to execute the move-to-counter action with the appropriate arguments. By doing this, the robot will move to the counter location and be ready to perform further actions to fulfil the goal.
\end{mdframed}

\section{Modular and Customizable Design} \label{sec:modular_design} \nameAbbr{} was consciously designed to be modular and customizable, allowing for easy upgrading or replacement of a number of components to suit particular educator needs. We aim to release \nameAbbr{} as an open source software like its predecessor, and hope that the community will build upon and improve our work in order to further the goal of making AI education more accessible to the general public. Despite the many improvements and feature additions in \nameAbbr{}, it is still a lightweight web-based interface which can run on any device with a modern browser. We have also containerized the software to allow the platform to be run on any machine without the user having to worry about installing dependencies. Containerization also assists in deployment over cloud based services allowing for convenient scaling as required, especially suited to schools and other educational settings.

\subsection{New Domains and Problems} \label{ssec:new_domains}Adding new domains simply requires the additional environment description dae files, the action configuration specifications of the domain, a semantic mapping of predicates and actions to natural language, and the domain and problem pddl files. \nameAbbr{} given all these inputs handles the creation of the blockly interface, web frontend, setting up the simulator as well as explanation generation on its own. Incorporating new problems within existing domains is even simpler and requires only the addition of new pddl problem files, or can be generated automatically using the curriculum generation module.

\subsection{LLMs} \label{ssec:llms_modular}There is flexibility in choosing the LLM used in translating explanations and hints to a user-friendly format. \nameAbbr{} can be used with any LLM, since the interface of passing the prompt abstracts away the internals of the explanation generation. The LLM being used may be stored locally in the system or accessed via an API. Educators, therefore, can use LLMs fine-tuned specifically for the task of converting predicates into human readable language, for instance. Further, the prompt being used to generate responses from the LLM can also be modified as required. different prompts can be used to generate responses in a specific format, or to be less or more verbose as needed.

\subsection{Hinting} \label{ssec:hinting_modular}Educators can make hints more, or less transparent to the students as they need. A single tunable real number parameter between 0 and 1 represents the independent probability of each grounded parameter in the action hint being displayed to the user. By increasing this value, hints are more likely to reveal the grounded parameters input to the action in the hint, and vice-versa.

\subsection{Simulator} \label{ssec:simulator_modular}In order to be simulator-agnostic, \nameAbbr{} separates the simulator from the rest of the backend software, and streams the output to the webpage. Our work uses OpenRave streamed using noVNC \cite{noVNC}, but any robot simulator package can be used in its place. 

\begin{table*}[t]    
    \centering
    \begin{tabular}{llccc}
         \toprule
        \textbf{Desiderata} & \textbf{\nameAbbr{}} & \textbf{JEDAI} & \textbf{Reference}  \\
         \midrule
            Open source & \cmark & \cmark  & \ref{sec:modular_design} \\
            Minimal system requirements & \cmark & \cmark  & \ref{sec:modular_design} \\
            Highly customizable & \cmark & \cmark  & \ref{sec:modular_design} \\
            Integrated simulation & \cmark & \cmark  & \ref{ssec:ui_improvements} \\            
            User-adaptive curriculum problem generation & \cmark & \xmark & \ref{ssec:curriculum_generation} \\                     
            LLM translated explanations & \cmark & \xmark  & \ref{ssec:intuitive_expl_hints}, \ref{ssec:llm_prompts} \\
            Hint messages & \cmark & \xmark  &  \ref{ssec:hinting_modular}\\
            LLM translated hint messages & \cmark & \xmark & \ref{ssec:intuitive_expl_hints} \\            
            State space annotation & \cmark & \xmark  & \ref{ssec:ss_display} \\
            Continuous plan checking and explanation generation & \cmark & \xmark &\ref{ssec:intuitive_ui}, \ref{ssec:ui_improvements}   \\            
            Explanation of multiple failing actions & \cmark & \xmark  & \ref{ssec:llm_prompts} \\            
            Collapsible information fields & \cmark & \xmark & \ref{ssec:intuitive_ui}, \ref{ssec:ui_improvements}  \\
            Minimize/Maximize blocks & \cmark & \xmark &  \ref{ssec:blockly_ui} \\
            Clicking on error message highlights corresponding action block  & \cmark & \xmark & \ref{ssec:ui_improvements}  \\
            Robot simulator in the same window & \cmark & \xmark & \ref{ssec:intuitive_ui}, \ref{ssec:ui_improvements}  \\
                        
         \bottomrule 
    \end{tabular}
    \caption{A comparison of the features of \nameAbbr{} compared to JEDAI.}
    \label{tab:desiderata_supplement}
\end{table*}

\newcolumntype{C}[1]{>{\arraybackslash}p{#1}}

\section{\nameAbbr{} Bundled Environment Details} \label{sec:bundled_environments}
Even though the educators can add custom environments to \nameAbbr{}, it comes preconfigured with a few environments to help educators. We have seen one such environment in Fig.~\ref{fig:screenshot} in the main paper: Fetch robot in the Coffee Shop environment task with delivering orders to various tables. We briefly introduce three more environments here:

\begin{figure}[th]
    \centering
    \includegraphics[width=\columnwidth]{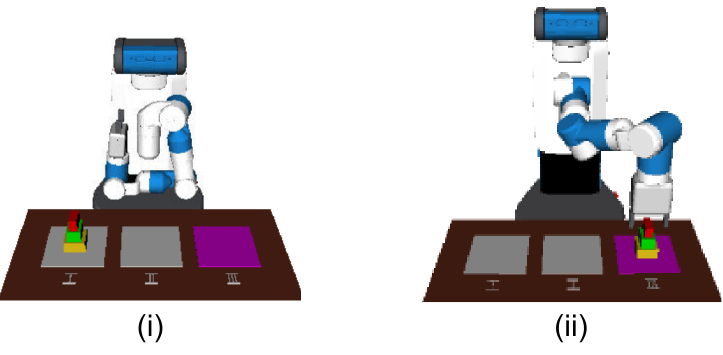}
    \caption{Screenshot of the simulator from the Tower of Hanoi environment. (i) depicts the initial state whereas (ii) a sample goal image for this environment.}
    \label{fig:hanoi}
\end{figure}

\paragraph{Tower of Hanoi} \label{ssec:hanoi}In this environment shown in Fig.~\ref{fig:hanoi}, Fetch robot has to solve the classical tower of Hanoi problem. It consists of three blocks on top of each other kept at a location. There are three such locations. All three blocks have to be transported to another fixed location, but the robot cannot place a larger block on top of a smaller block. The blocks are uniquely identified using their size.

\begin{figure}[th]
    \centering
    \includegraphics[width=\columnwidth]{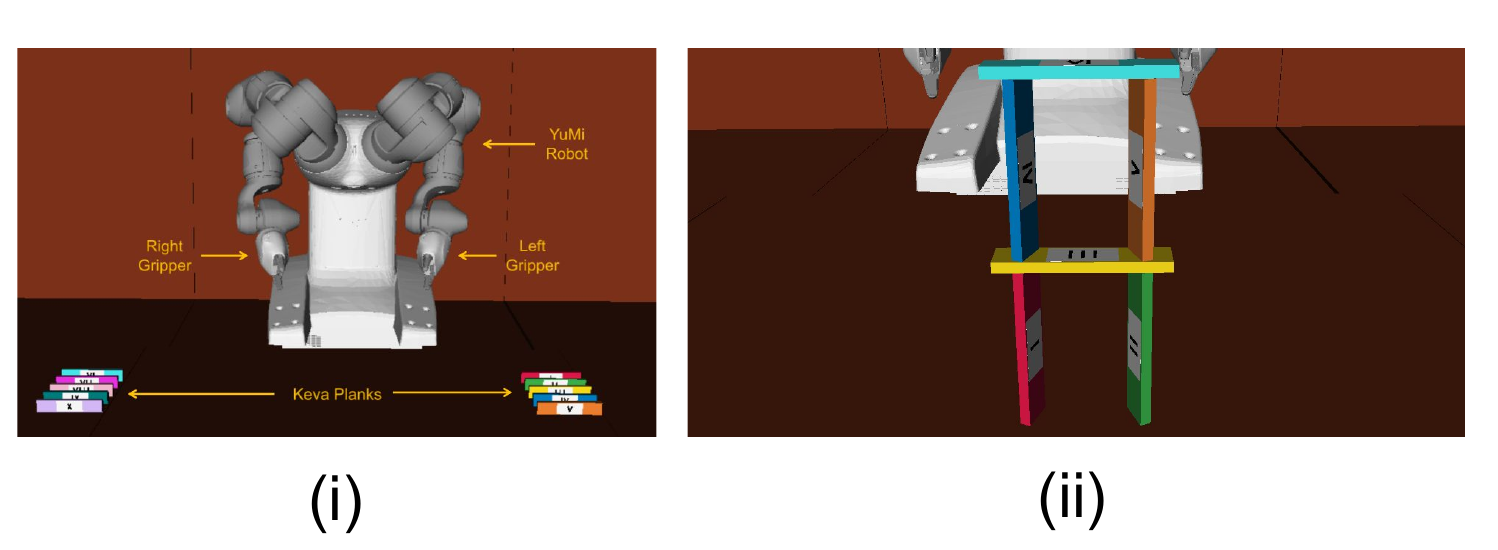}
    \caption{Screenshot of the simulator from the keva environment. (i) depicts the initial state whereas (ii) a sample goal image for this environment.}
    \label{fig:keva}
\end{figure}

\paragraph{Keva Planks} \label{ssec:keva}In this environment shown in Fig.~\ref{fig:keva}, YuMi, a dual-armed robot has to create structures using numbered planks kept on the table. Different structures can be created by placing planks vertically, horizontally or along their edges. Planks can be placed on the table, or on top of other planks.

\begin{figure}[th]
    \centering
    \includegraphics[width=\columnwidth]{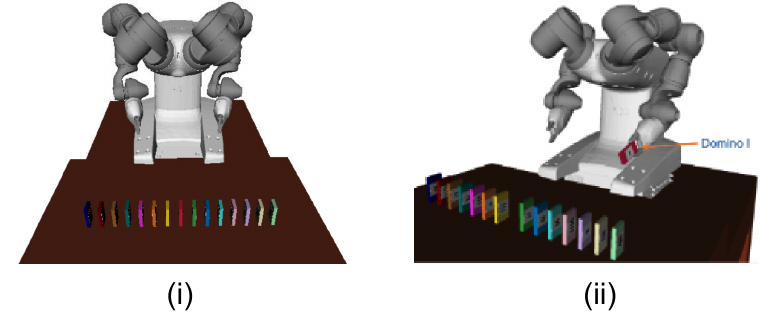}
    \caption{Screenshot of the simulator from the dominoes environment. (i) depicts the initial state whereas (ii) a sample goal image for this environment.}
    \label{fig:dominoes}
\end{figure}

\paragraph{Dominoes} \label{ssec:dominoes}In this environment shown in Fig.~\ref{fig:dominoes}, similar to the Keva planks setup, YuMi robot has to create structures using dominoes. Each domino can be uniquely identified using an ID that is printed on each domino. The users can move around the camera in the simulator to view these IDs.

\section{Study Details}\label{sec:study_details}
As mentioned in the main paper, the study was divided into four phases. This section provides complete data for the results (Fig.\,\ref{fig:user_study_results}, Table\,\ref{fig:survey}) presented in the main paper. 

\subsection{Pre-survey Phase}\label{ssec:pre_survey}
Table\,\ref{tab:pre} shows the questions and collected responses from the pre-survey questionnaire that was administered as a part of this phase. Note that the participants were yet to interact with \nameAbbr{} or JEDAI in this phase. We provide the statistical significance test results from both the overall participant pool and also segregate the responses of the participants who were assigned \nameAbbr{} or JEDAI in the next phase. Our results (one-sample t-test) show that participant responses are statistically significant across these dimensions ($\mu^1_0=2$) except for Pre Q5. For Pre Q5, both the combined and per-system statistical results are insignificant. Thus, participants found it neither easy nor difficult to program robots. Similarly, we cannot reject the null hypothesis from the results of the two-sample t-test (that compares responses between \nameAbbr{} and JEDAI for the questions) thus showing that participants assigned to either \nameAbbr{} or JEDAI had similar demographics in our study.

Our pre-study results also show that participants had little to no experience with robotics programming (Pre Q6) and $\ge 50\%$ think about AI (Pre Q2) or use AI-based tools (Pre Q3) daily. This indicates the greater need to impart AI education to people who will be interacting with AI systems frequently.

\subsection{Training Phase}\label{ssec:training}
There was no questionnaire administered to the participants at the end of this phase.
\subsection{Test Phase}\label{ssec:test}
Table\,\ref{tab:post} and Table\,\ref{tab:post2} present the results obtained after administering the post-survey questionnaire at the end of this phase. Depending upon the control group to which they were assigned, participants had interacted with \nameAbbr{} or JEDAI (but not both) at this point. We report these results pictorially in Fig.\,\ref{fig:user_study_results} of the main paper. Post Q1, Q3, Q9, and Q10 are not discussed in the main paper. The results for Post Q1 are not statistically significant. This is not surprising since both \nameAbbr{} and JEDAI manage to increase user understanding of the robot's limitations and capabilities (Post Q8, H6).  We also report results on performing single-tailed two-sample t-tests to analyze responses across the control groups. We use the null hypotheses of no difference ($\mu = 0$) to analyze these results and focus on only a single tail that allows us to analyze whether users prefer \nameAbbr{} over JEDAI. We report the p-values obtained from these tests. Statistically significant responses ($p < 0.05$) indicate that users preferred \nameAbbr{} over JEDAI in their response.

Table\,\ref{tab:post2} contains \nameAbbr{} specific questions (Q3, Q9, Q10) meant to evaluate the results of the hinting feature of \nameAbbr{}. We did not administer these questions to participants interacting with JEDAI since IRB protocol forbids the exposure of questions that can allow users to infer the presence of a different control group. Our results show that users the test problem moderately challenging and utilized hints only a few times. We attribute this to the training phase where the adaptive task generation algorithm taught the users appropriately such that their need for hints was reduced. Since there was no study administered during the training phase we do not have any data pertaining to the usage of hints during training.

In Table\,\ref{tab:post}, Post Q2 and in Table\,\ref{tab:post2} Post Q10 provided students with a sixth option for the users to choose - "I did not encounter any error" and "I did not receive any "Hints"" respectively. These options were provided to handle the edge cases that the users might experience where for Post Q2, the user do not make a mistake while completing the given problem and for Post Q10, the user did not use the "Get A Hint" button of the interface to get a hint for next step. In Table\,\ref{tab:post}, there is one student who chose the sixth option in Post Q2 after interaction with JEDAI as the first system. In Table\,\ref{tab:post2}, seven users opted for sixth option in Post Q10 after interacting with \nameAbbr{} as their first system. These data points for the respective questions were not considered for the statistical analysis of the aforementioned questions because this options do not adhere to the Likert scale that we used for the analysis.

\subsection{Sentiment Change Phase}\label{ssec:sentiment_change}

Table\,\ref{tab:sent_change1} and Table\,\ref{tab:sent_change2} presents the overall Sentiment Change that we observed from the data collected from the user study. These results were obtained after the users were asked to interact with second system($S_2$). Table\,\ref{tab:sent_change1} contains the overall sentiment change (both positive and negative) for users who were given \nameAbbr{} as their first system($S_1$) and JEDAI as second system($S_2$). Table\,\ref{tab:sent_change2} contains the results for sentiment change (both positive and negative) for users who interacted with JEDAI first and then 
with \nameAbbr{}. 

We don't report the paired t-test values for these groups because for most of the questions, the paired distribution does not satisfy the normality assumption of paired t-test. We tested the normality of the data by performing d'Agostino-Pearson test \cite{pearsonPtest}. We were unable to run Wilcoxon Signed-Rank \cite{wilcoxonSignedTest} test because the paired data has too many ties and the sample size was insufficient to run the test. 

As discussed in section \ref{ssec:test}, Post Q2 in Table\,\ref{tab:sent_change1} and Table\,\ref{tab:sent_change2} have sixth option provided to the users that does not adhere to Likert scale. Also, as per IRB protocol, users can choose not to respond to any of the questions in the survey. For both some questions in both the mentioned tables, there are users who have either opted for option sixth in Post Q2 or have chosen not to respond to some questions. In both the cases, we did not consider the data points where either the user has opted for the sixth option or has chosen not to respond as both scenarios differ from Likert scale options that we have chosen for the analysis of the questions. 
In Table\,\ref{tab:sent_change1}, for JEDAI - Post Q1, 1 user has no response and for JEDAI Post Q2, 4 users have opted for sixth option. In Table\,\ref{tab:sent_change2}, for \nameAbbr{} Post Q2, 2 users have opted for sixth option.

\begin{table*}
\small
\begin{tabularx}{2\columnwidth}{p{1.1cm}p{1.2cm}p{1.25cm}p{1.6cm}p{1.4cm}p{1.2cm}ccp{1.2cm}cc}
\toprule
& \multicolumn{5}{c}{} & \multicolumn{5}{c}{t-tests} \\
\cmidrule{7-11}
& \multicolumn{5}{c}{} & \multicolumn{3}{c}{One-sample} & & \multicolumn{1}{c}{Two-sample} \\
\cmidrule{7-9}  \cmidrule{11-11}
Question & \multicolumn{5}{c}{Question and Responses} & \multicolumn{1}{c}{$\mu_0^1$} & 
\multicolumn{1}{c}{$\mu$} & 
\multicolumn{1}{c}{$p$} & 
& 
\multicolumn{1}{c}{$p$} \\

\midrule

Pre Q1 & \multicolumn{8}{p{8.5cm}}{
How familiar are you with Computer Science and Artificial Intelligence (A.I.)? 
} \\
& 
\makecell[lt]{Not at \\ all (0)} \CC &
Slightly well (1) \CC &
Moderately well (2) \CC &
\makecell[lt]{Very well \\(3)} \CC &
Extremely well (4) \CC \\

Total & 3 & 14 & 18 & 6 & 1 & \multirow{3}{*}{0} & 1.71 $\pm$ 0.89 & 1.56e-15 & & 0.36e-00  \\
\nameAbbr{} & 3 & 5 & 8 & 4 & 1 & & 1.76 $\pm$ 1.09 & 3.80e-07  \\
JEDAI & 0 & 9 & 10 & 2 & 0 & & 1.66 $\pm$ 0.65 & 2.46e-10  \\
\midrule

Pre Q2 & \multicolumn{8}{p{8.5cm}}{
How often do you think about A.I. in your day-to-day life?? 
} \\
& 
\makecell[lt]{Never (0)} \CC &
\makecell[lt]{Once a \\ week (1)} \CC &
2-3 times a week (2) \CC &
4-6 times a week (3) \CC &
Daily (4) \CC \\

Total & 1 & 3 & 8 & 11 & 19 & \multirow{3}{*}{2} & 3.04 $\pm$ 1.08 & 8.61e-08 & & 0.58e-01  \\
\nameAbbr{} & 1 & 0 & 3 & 4 & 13 & & 3.33 $\pm$ 1.06 & 6.44e-06 \\
JEDAI & 0 & 3 & 5 & 7 & 6 & & 2.76 $\pm$ 1.044 & 0.16e-02 \\
\midrule

Pre Q3 & \multicolumn{8}{p{8.5cm}}{
How often do you interact with tools that use A.I.?
} \\
& 
\makecell[lt]{Never (0)} \CC &
\makecell[lt]{Once a \\ week (1)} \CC &
2-3 times a week (2) \CC &
4-6 times a week (3) \CC &
Daily (4) \CC \\

Total & 2 & 4 & 5 & 10 & 21 & \multirow{3}{*}{1} & 3.04 $\pm$ 1.20 & 4.41e-14 & & 0.81e-01 \\
\nameAbbr{} & 0 & 2 & 2 & 4 & 13 & & 3.33 $\pm$ 1.01 & 6.71e-10  \\
JEDAI & 2 & 2 & 3 & 6 & 8 & & 2.76 $\pm$ 1.33 & 3.36e-06  \\
\midrule

Pre Q4 & \multicolumn{8}{p{8.5cm}}{
How curious are you to learn about the extent to which A.I. systems and robots can be used today? 
} \\
& 
\makecell[lt]{Not \\curious (0)} \CC &
\makecell[lt]{Slightly \\ curious (1)} \CC &
Moderately curious (2) \CC &
\makecell[lt]{Very \\ curious (3)}\CC &
\makecell[lt]{Extremely \\ curious (4)} \CC \\

Total & 1 & 4 & 3 & 21 & 13 & \multirow{3}{*}{2} & 2.97 $\pm$ 0.99 & 1.47e-07 & & 0.35e-00 \\
\nameAbbr{} & 0 & 2 & 1 & 12 & 6 & & 3.04 $\pm$ 0.86 & 1.95e-05  \\
JEDAI & 1 & 2 & 2 & 9 & 7 & & 2.90 $\pm$ 1.13 & 0.15e-02  \\
\midrule

Pre Q5 & \multicolumn{8}{p{8.5cm}}{
Assume you want a household robot to get you water from the refrigerator. How difficult do you think it is to give it instructions to perform this task?  
} \\
& 
\makecell[lt]{Very \\ difficult (0)} \CC &
\makecell[lt]{Slightly \\difficult (1)} \CC &
Neither difficult nor easy (2) \CC &
Slightly easy (3) \CC &
Very easy (4) \CC \\

Total & 3 & 20 & 6 & 10 & 3 & \multirow{3}{*}{2} & 2.23 $\pm$ 1.12 & 0.17e-00 & & 0.93e-01  \\
\nameAbbr{} & 2 & 7 & 4 & 5 & 3 & & 2.00 $\pm$ 1.26 & 1.00e-00  \\
JEDAI & 1 & 13 & 2 & 5 & 0 & & 2.47 $\pm$ 0.92 & 0.29e-01 \\
\midrule

Pre Q6 & \multicolumn{8}{p{8.5cm}}{
How familiar are you with robotics programming?
} \\
& 
Not at all (0) \CC &
\makecell[lt]{Slightly \\familiar (1)}\CC &
Moderately familiar (2) \CC &
\makecell[lt]{Very\\ familiar (3)} \CC &
\makecell[lt]{Extremely \\familiar (4)} \CC \\

Total & 26 & 13 & 3 & 0 & 0 & \multirow{3}{*}{0} & 0.45 $\pm$ 0.63 & 3.59e-05 & & 0.26e-00 \\
\nameAbbr{} & 11 & 9 & 1 & 0 & 0 & & 0.52 $\pm$ 0.60 & 0.71e-03 \\
JEDAI & 15 & 4 & 2 & 0 & 0 & & 0.38 $\pm$ 0.66 & 0.16e-01 \\

  \bottomrule
\end{tabularx}
\caption{Pre-survey questionnaire administered during the \emph{pre-survey phase} of our user study ($\mu^1_0$ represents the null hypothesis mean used to conduct the one-sample t-test). We used $\alpha=0.05$ for determining statistical significance for the t-tests. For each Question ID, the first row is the question as it was presented to the participants. The second row lists the possible answers for the question (and the corresponding Likert scale values in parentheses). The third row presents the responses by all participants. The breakdown of the total responses by control group are presented in the fourth and fifth rows.}
\label{tab:pre}
\end{table*}

\begin{table*}
\small
\begin{tabularx}{2\columnwidth}{p{1.15cm}p{1.5cm}p{1.5cm}p{1.6cm}p{1.4cm}p{1.55cm}ccp{0.0cm}c}
\toprule
& \multicolumn{5}{c}{} & \multicolumn{4}{c}{t-tests} \\
\cmidrule{7-10}
& \multicolumn{5}{c}{} & \multicolumn{2}{c}{One-sample} & & \multicolumn{1}{c}{Two-sample} \\
\cmidrule{7-8}  \cmidrule{10-10}

Question & \multicolumn{5}{c}{Question and Responses} & $\mu$ & $p$ & & $p$ \\
\midrule
      Post Q1 & \multicolumn{8}{p{9.5cm}}{After interacting with the \nameAbbr{} system, how inclined are you to learn how daily problems are being solved with A.I.? }\\
       & \makecell[lt]{Not \\inclined (0)} \CC 
       & \makecell[lt]{Slightly \\inclined (1)}\CC
       & \makecell[lt]{Moderately \\inclined (2)}\CC
       & \makecell[lt]{Very \\inclined (3)} \CC 
       & \makecell[lt]{Extremely \\inclined (4)}\CC
       \\
       {\nameAbbr{}} & 0 & 2 & 6 & 6 & 7 & 2.85 $\pm$ 1.01 & 5.52e-08 & & \multirow{2}{*}{0.14e-00}\\
        JEDAI & 2 & 3 & 4 & 11 & 1 & 2.28 $\pm$ 1.10 & 3.11e-05\\
       \midrule
     Post Q2 & \multicolumn{8}{p{9.5cm}}{How helpful were the explanations that were given for the cause of an error? }\\
       & \makecell[lt]{Not \\helpful (0)}\CC 
       & \makecell[lt]{Slightly \\helpful (1)} \CC 
       & Moderately helpful (2)\CC 
       & \makecell[lt]{Very \\helpful (3)}\CC 
       & \makecell[lt]{Extremely \\helpful (4)}\CC 
       \\
       {\nameAbbr{}} & 0 & 1 & 2 & 6 & 12 & 3.38 $\pm$ 0.86 & 2.23e-07 & & \multirow{2}{*}{0.67e-02} \\
       JEDAI & 1 & 4 & 5 & 5 & 5 & 2.42 $\pm$ 1.20 & 0.59e-01 \\
       \midrule
       Post Q4 & \multicolumn{8}{p{9.5cm}}{How intuitive was the interface? }\\
       & \makecell[lt]{Not \\intuitive (0)}\CC 
       & \makecell[lt]{Slightly \\intuitive (1)}\CC 
       & \makecell[lt]{Moderately \\intuitive (2)}\CC 
       & \makecell[lt]{Very \\intuitive (3)}\CC 
       & \makecell[lt]{Extremely \\intuitive (4)} \CC 
       \\
       {\nameAbbr{}} & 0 & 1 & 6 & 12 & 2 & 2.27 $\pm$ 0.71 & 9.41e-05 & & \multirow{2}{*}{0.22e-01}\\
       JEDAI & 0 & 4 & 11 & 4 & 2 & 2.19 $\pm$ 0.87 & 0.16e-00\\
       \midrule
        Post Q5 & \multicolumn{8}{p{9.5cm}}{As compared to before participating in this user study, how much has your curiosity increased to learn more about AI systems and robots? }\\
       & \makecell[lt]{Highly \\decreased (0)} \CC 
       & \makecell[lt]{Slightly \\decreased (1)}\CC 
       & \makecell[lt]{Neither \\increased nor \\decreased (2)}\CC 
       & \makecell[lt]{Slightly \\increased (3)}\CC 
       & \makecell[lt]{Highly \\increased (4)}\CC 
       \\
       {\nameAbbr{}} & 0 & 0 & 1 & 9 & 11 & 3.47 $\pm$ 0.60 & 4.24e-10 & & \multirow{2}{*}{0.28e-01}\\
       JEDAI & 0 & 1 & 5 & 8 & 7 & 3.00 $\pm$ 0.89 & 5.17e-05\\
       \midrule
       Post Q6 & \multicolumn{8}{p{9.5cm}}{How well do you think you now understand how one can use an AI system to make a plan for a robot to perform a task? }\\
       & \makecell[lt]{Not well (0)} \CC 
       & Slightly well (1)\CC 
       & Moderately well (2)\CC 
       & Very well (3)\CC 
       & Extremely well (4)\CC 
       \\
       {\nameAbbr{}} & 0 & 1 & 5 & 11 & 4 & 2.85 $\pm$ 0.79 & 3.81e-05 & & \multirow{2}{*}{0.46e-01}\\
       JEDAI & 1 & 4 & 5 & 9 & 2 & 2.33 $\pm$ 1.06 & 0.83e-01\\
       \midrule
       Post Q7 & \multicolumn{8}{p{9.5cm}}{Do you agree that the \nameAbbr{} system made it easier for you to provide instructions to a robot for performing tasks? }\\
       & \makecell[lt]{Strongly \\Disagree (0)}\CC 
       & Disagree (1)\CC 
       & \makecell[lt]{Neither \\Agree nor \\Disagree (2)} \CC 
       & Agree (3) \CC 
       & \makecell[lt]{Strongly \\Agree (4)}\CC 
       \\
       {\nameAbbr{}} & 0 & 0 & 1 & 9 & 11 & 3.47 $\pm$ 0.60 & 4.24e-10 & & \multirow{2}{*}{0.23e-01}\\
       JEDAI & 0 & 1 & 3 & 11 & 6 & 3.04 $\pm$ 0.80 & 7.81e-06\\
       \midrule
       Post Q8 & \multicolumn{8}{p{9.5cm}}{Do you agree that \nameAbbr{} helps improve the understanding of the robot’s limitations and capabilities? }\\
       & \makecell[lt]{Strongly \\Disagree (0)}\CC 
       & Disagree (1)\CC 
       & \makecell[lt]{Neither \\Agree nor \\Disagree (2)} \CC 
       & Agree (3) \CC 
       & \makecell[lt]{Strongly \\Agree (4)}\CC
       \\
       {\nameAbbr{}} & 0 & 0 & 2 & 12 & 7 & 3.23 $\pm$ 0.62 & 1.56e-08 & & \multirow{2}{*}{0.64e-01}\\
       JEDAI & 0 & 1 & 4 & 12 & 4 & 2.90 $\pm$ 0.76 & 2.78e-05\\
       \\
       \bottomrule

    \end{tabularx}
        \caption{Post-survey questionnaire administered during the \emph{test phase} of our user study ($\mu^1_0 = 2$ for the one-sample t-test). We used $\alpha=0.05$ for determining statistical significance for the t-tests. For each Question ID, the first row is the question as it was presented to the participants. The second row lists the possible answers for the question (and the corresponding Likert scale values in parentheses). Third and fourth row represent the number of participants who chose that answer for the question after \nameAbbr{} and JEDAI respectively. We include here only those questions that were presented to both control groups.}
    \label{tab:post}
\end{table*}

\begin{table*}
\small
\begin{tabularx}{2\columnwidth}{p{1.15cm}p{1.8cm}p{1.9cm}p{1.8cm}p{1.8cm}p{2.2cm}ccp{0.0cm}c}
\toprule
& \multicolumn{5}{c}{} & \multicolumn{2}{c}{One-sample t-test} \\
\cmidrule{7-8} 

Question & \multicolumn{5}{c}{Question and Responses} & $\mu$ & $p$ & \\
\midrule
Post Q3 & \multicolumn{8}{p{11cm}}{How challenging were the problems in the \nameAbbr{} session? }\\
       & \makecell[lt]{Not \\challenging (0)} \CC 
       & \makecell[lt]{Slightly \\challenging (1)}\CC 
       & \makecell[lt]{Moderately \\challenging (2)}\CC 
       & \makecell[lt]{Very \\challenging (3)}\CC 
       & \makecell[lt]{Extremely \\challenging (4)}\CC 
       \\
       {\nameAbbr{}} & 3 & 10 & 6 & 1 & 1 & 1.38 $\pm$ 0.97 & 0.88e-01 \\
       \midrule
       Post Q9 & \multicolumn{8}{p{11cm}}{How often did you use the “Hint” button during the hands-on session? }\\
       & \makecell[lt]{Once per \\problem (0)}\CC 
       & \makecell[lt]{2-4 times \\per problem (1)}\CC 
       & \makecell[lt]{Once during \\the session (2)}\CC 
       & Never (3) \CC 
       & \makecell[lt]{Didn't notice any \\hint button (4)} \CC 
       \\
       {\nameAbbr{}} & 4 & 3 & 9 & 5 & 0 & 2.23 $\pm$ 0.99 & 0.28e-00 \\
       \midrule
       Post Q10 & \multicolumn{8}{p{11cm}}{How well do you think “Hint” functionality helped you while solving the problems? }\\
       & \makecell[lt]{Not well (0)} \CC 
       & \makecell[lt]{Slightly well (1)}\CC 
       & \makecell[lt]{Moderately \\ well (2)}\CC 
       & \makecell[lt]{Very well (3)}\CC 
       & \makecell[lt]{Extremely well (4)}\CC 
       \\
       {\nameAbbr{}} & 0 & 4 & 5 & 3 & 2 & 2.21 $\pm$ 1.05 & 0.45e-00 \\
       \bottomrule
\end{tabularx}
        \caption{Post-survey questionnaire administered during the \emph{test phase} of our user study ($\mu^1_0 = 2$ for the one-sample t-test). We used $\alpha=0.05$ for determining statistical significance for the t-tests. For each Question ID, the first row is the question as it was presented to the participants. The second row lists the possible answers for the question (and the corresponding Likert scale values in parentheses). The third row represents the number of participants who chose that answer for the question after interacting with \nameAbbr{}. Since JEDAI does not include hints, per IRB protocol, these questions were excluded from users who interacted with JEDAI immediately before being administered the survey (elaborated in App.\,\ref{ssec:test}).}
\label{tab:post2}
\end{table*}

\begin{table*}
\small
\begin{tabularx}{2\columnwidth}{p{1.8cm}p{1.5cm}p{1.5cm}p{1.6cm}p{1.4cm}p{1.55cm}ccc}
\toprule
& \multicolumn{5}{c}{} & \multicolumn{2}{c}{Total Sentiment} \\
\cmidrule{7-8}

Question & \multicolumn{5}{c}{Question and Responses} & \multicolumn{1}{c}{Positive} & \multicolumn{1}{c}{Negative} \\
\midrule
      Post Q1 & \multicolumn{8}{p{9.5cm}}{After interacting with the \nameAbbr{} system, how inclined are you to learn how daily problems are being solved with A.I.? }\\
       & \makecell[lt]{Not \\inclined (0)} \CC 
       & \makecell[lt]{Slightly \\inclined (1)}\CC
       & \makecell[lt]{Moderately \\inclined (2)}\CC
       & \makecell[lt]{Very \\inclined (3)} \CC 
       & \makecell[lt]{Extremely \\inclined (4)}\CC
       \\
       $S_1$=\nameAbbr{} & 0 & 2 & 6 & 6 & 7 & 13 & 2 & \\
       $S_2$=JEDAI & 0 & 5 & 5 & 3 & 7 & 10 & 5 & \\
       \midrule
     Post Q2 & \multicolumn{8}{p{9.5cm}}{How helpful were the explanations that were given for the cause of an error? }\\
       & \makecell[lt]{Not \\helpful (0)}\CC 
       & \makecell[lt]{Slightly \\helpful (1)} \CC 
       & Moderately helpful (2)\CC 
       & \makecell[lt]{Very \\helpful (3)}\CC 
       & \makecell[lt]{Extremely \\helpful (4)}\CC 
       \\
       $S_1$=\nameAbbr{} & 0 & 1 & 2 & 6 & 12 & 18 & 1 & \\
       $S_2$=JEDAI & 0 & 3 & 3 & 5 & 6 & 11 & 3 &\\
       \midrule
       Post Q4 & \multicolumn{8}{p{9.5cm}}{How intuitive was the interface? }\\
       & \makecell[lt]{Not \\intuitive (0)}\CC 
       & \makecell[lt]{Slightly \\intuitive (1)}\CC 
       & \makecell[lt]{Moderately \\intuitive (2)}\CC 
       & \makecell[lt]{Very \\intuitive (3)}\CC 
       & \makecell[lt]{Extremely \\intuitive (4)} \CC 
       \\
       $S_1$=\nameAbbr{} & 0 & 1 & 6 & 12 & 2 & 14 & 1 & \\
       $S_2$=JEDAI & 0 & 1 & 11 & 7 & 2 & 9 & 1 & \\
       \midrule
        Post Q5 & \multicolumn{8}{p{9.5cm}}{As compared to before participating in this user study, how much has your curiosity increased to learn more about AI systems and robots? }\\
       & \makecell[lt]{Highly \\decreased (0)} \CC 
       & \makecell[lt]{Slightly \\decreased (1)}\CC 
       & \makecell[lt]{Neither \\increased nor \\decreased (2)}\CC 
       & \makecell[lt]{Slightly \\increased (3)}\CC 
       & \makecell[lt]{Highly \\increased (4)}\CC 
       \\
       $S_1$=\nameAbbr{} & 0 & 0 & 1 & 9 & 11 & 20 & 0 & \\
       $S_2$=JEDAI & 0 & 0 & 1 & 11 & 9 & 20 & 0 & \\
       \midrule
       Post Q6 & \multicolumn{8}{p{9.5cm}}{How well do you think you now understand how one can use an AI system to make a plan for a robot to perform a task? }\\
       & \makecell[lt]{Not well (0)} \CC 
       & Slightly well (1)\CC 
       & Moderately well (2)\CC 
       & Very well (3)\CC 
       & Extremely well (4)\CC 
       \\
       $S_1$=\nameAbbr{} & 0 & 1 & 5 & 11 & 4 & 15 & 1 & \\
       $S_2$=JEDAI & 1 & 4 & 4 & 8 & 4 & 12 & 5 &\\
       \midrule
       Post Q7 & \multicolumn{8}{p{9.5cm}}{Do you agree that the \nameAbbr{} system made it easier for you to provide instructions to a robot for performing tasks? }\\
       & \makecell[lt]{Strongly \\Disagree (0)}\CC 
       & Disagree (1)\CC 
       & \makecell[lt]{Neither \\Agree nor \\Disagree (2)} \CC 
       & Agree (3) \CC 
       & \makecell[lt]{Strongly \\Agree (4)}\CC 
       \\
       $S_1$=\nameAbbr{} & 0 & 0 & 1 & 9 & 11 & 20 & 0 &\\
       $S_2$=JEDAI & 1 & 0 & 2 & 11 & 7 & 18 & 1 & \\
       \midrule
       Post Q8 & \multicolumn{8}{p{9.5cm}}{Do you agree that \nameAbbr{} helps improve the understanding of the robot’s limitations and capabilities? }\\
       & \makecell[lt]{Strongly \\Disagree (0)}\CC 
       & Disagree (1)\CC 
       & \makecell[lt]{Neither \\Agree nor \\Disagree (2)} \CC 
       & Agree (3) \CC 
       & \makecell[lt]{Strongly \\Agree (4)}\CC
       \\
       $S_1$=\nameAbbr{} & 0 & 0 & 2 & 12 & 7 & 19 & 0 & \\
       $S_2$=JEDAI & 1 & 0 & 0 & 13 & 7 & 20 & 1 & \\
       \bottomrule 
       
    \end{tabularx}
        \caption{Post-survey questionnaire administered during the \emph{sentiment change phase} of our user study. For each Question ID, the first row is the question as it was presented to the participants. The second row lists the possible answers for the question (and the corresponding Likert scale values in parentheses). Third and fourth row represent the number of participants who chose that answer for the question after interacting with \nameAbbr{} and JEDAI respectively.}    
    \label{tab:sent_change1}
\end{table*}

\begin{table*}
\small
\begin{tabularx}{2\columnwidth}{p{1.8cm}p{1.5cm}p{1.5cm}p{1.6cm}p{1.4cm}p{1.55cm}ccc}
\toprule
& \multicolumn{5}{c}{} & \multicolumn{2}{c}{Total Sentiment} \\
\cmidrule{7-8}

Question & \multicolumn{5}{c}{Question and Responses} & \multicolumn{1}{c}{Positive} & \multicolumn{1}{c}{Negative} \\
\midrule
      Post Q1 & \multicolumn{8}{p{9.5cm}}{After interacting with the \nameAbbr{} system, how inclined are you to learn how daily problems are being solved with A.I.? }\\
       & \makecell[lt]{Not \\inclined (0)} \CC 
       & \makecell[lt]{Slightly \\inclined (1)}\CC
       & \makecell[lt]{Moderately \\inclined (2)}\CC
       & \makecell[lt]{Very \\inclined (3)} \CC 
       & \makecell[lt]{Extremely \\inclined (4)}\CC
       \\
        $S_1$=JEDAI & 2 & 3 & 4 & 11 & 1 & 12 & 5 & \\
       $S_2$=\nameAbbr{} & 0 & 3 & 6 & 10 & 2 & 12 & 3 & \\
       \midrule
     Post Q2 & \multicolumn{8}{p{9.5cm}}{How helpful were the explanations that were given for the cause of an error? }\\
       & \makecell[lt]{Not \\helpful (0)}\CC 
       & \makecell[lt]{Slightly \\helpful (1)} \CC 
       & Moderately helpful (2)\CC 
       & \makecell[lt]{Very \\helpful (3)}\CC 
       & \makecell[lt]{Extremely \\helpful (4)}\CC 
       \\
       $S_1$=JEDAI & 1 & 4 & 5 & 5 & 5 & 10 & 5 &  \\
       $S_2$=\nameAbbr{} & 0 & 1 & 2 & 8 & 8 & 16 & 1 &  \\
       \midrule
       Post Q4 & \multicolumn{8}{p{9.5cm}}{How intuitive was the interface? }\\
       & \makecell[lt]{Not \\intuitive (0)}\CC 
       & \makecell[lt]{Slightly \\intuitive (1)}\CC 
       & \makecell[lt]{Moderately \\intuitive (2)}\CC 
       & \makecell[lt]{Very \\intuitive (3)}\CC 
       & \makecell[lt]{Extremely \\intuitive (4)} \CC 
       \\
       $S_1$=JEDAI & 0 & 4 & 11 & 4 & 2 & 6 & 4 & \\
       $S_2$=\nameAbbr{} & 0 & 3 & 6 & 8 & 4 & 12 & 3 & \\
       \midrule
        Post Q5 & \multicolumn{8}{p{9.5cm}}{As compared to before participating in this user study, how much has your curiosity increased to learn more about AI systems and robots? }\\
       & \makecell[lt]{Highly \\decreased (0)} \CC 
       & \makecell[lt]{Slightly \\decreased (1)}\CC 
       & \makecell[lt]{Neither \\increased nor \\decreased (2)}\CC 
       & \makecell[lt]{Slightly \\increased (3)}\CC 
       & \makecell[lt]{Highly \\increased (4)}\CC 
       \\
       $S_1$=JEDAI & 0 & 1 & 5 & 8 & 7 & 12 & 1 & \\
       $S_2$=\nameAbbr{} & 0 & 0 & 4 & 11 & 6 & 17 & 0 & \\
       \midrule
       Post Q6 & \multicolumn{8}{p{9.5cm}}{How well do you think you now understand how one can use an AI system to make a plan for a robot to perform a task? }\\
       & \makecell[lt]{Not well (0)} \CC 
       & Slightly well (1)\CC 
       & Moderately well (2)\CC 
       & Very well (3)\CC 
       & Extremely well (4)\CC 
       \\
       $S_1$=JEDAI & 1 & 4 & 5 & 9 & 2 & 11 & 5 & \\
       $S_2$=\nameAbbr{} & 0 & 4 & 6 & 9 & 2 & 11 & 4 & \\
       \midrule
       Post Q7 & \multicolumn{8}{p{9.5cm}}{Do you agree that the \nameAbbr{} system made it easier for you to provide instructions to a robot for performing tasks? }\\
       & \makecell[lt]{Strongly \\Disagree (0)}\CC 
       & Disagree (1)\CC 
       & \makecell[lt]{Neither \\Agree nor \\Disagree (2)} \CC 
       & Agree (3) \CC 
       & \makecell[lt]{Strongly \\Agree (4)}\CC 
       \\
       $S_1$=JEDAI & 0 & 1 & 3 & 11 & 6 & 17 & 1 & \\
       $S_2$=\nameAbbr{} & 0 & 0 & 1 & 14 & 6 & 20 & 0 & \\
       \midrule
       Post Q8 & \multicolumn{8}{p{9.5cm}}{Do you agree that \nameAbbr{} helps improve the understanding of the robot’s limitations and capabilities? }\\
       & \makecell[lt]{Strongly \\Disagree (0)}\CC 
       & Disagree (1)\CC 
       & \makecell[lt]{Neither \\Agree nor \\Disagree (2)} \CC 
       & Agree (3) \CC 
       & \makecell[lt]{Strongly \\Agree (4)}\CC
       \\
       $S_1$=JEDAI & 0 & 1 & 4 & 12 & 4 & 16 & 1 & \\
       $S_2$=\nameAbbr{} & 0 & 0 & 5 & 11 & 5 & 16 & 0 & \\
       \bottomrule 
       
    \end{tabularx}
        \caption{Post-survey questionnaire administered during the \emph{sentiment change phase} of our user study. For each Question ID, the first row is the question as it was presented to the participants. The second row lists the possible answers for the question (and the corresponding Likert scale values in parentheses). Third and fourth row represent the number of participants who chose that answer for the question after interacting with \nameAbbr{} and JEDAI respectively.}    
    \label{tab:sent_change2}
\end{table*}

\clearpage

\onecolumn

\begin{landscape}
    
    \begin{figure}[ht]
        \centering
        \includegraphics[width=\columnwidth]{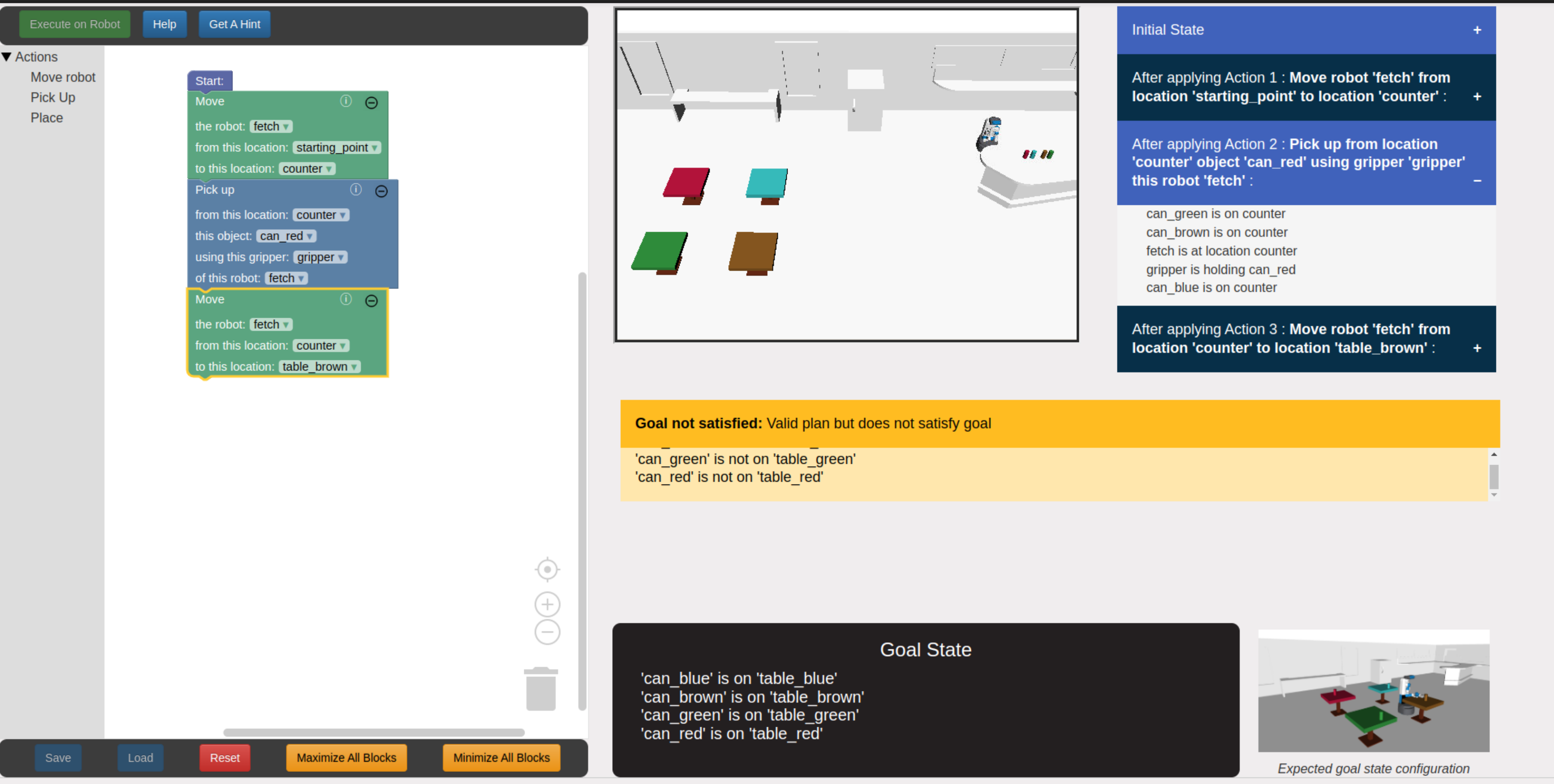}
    
        \caption{Screenshot of the \nameAbbr{} user interface.}
        \label{fig:unmod_jedai_ed}
    \end{figure}

    \begin{figure}[ht]
        \centering
            \includegraphics[width=\columnwidth]{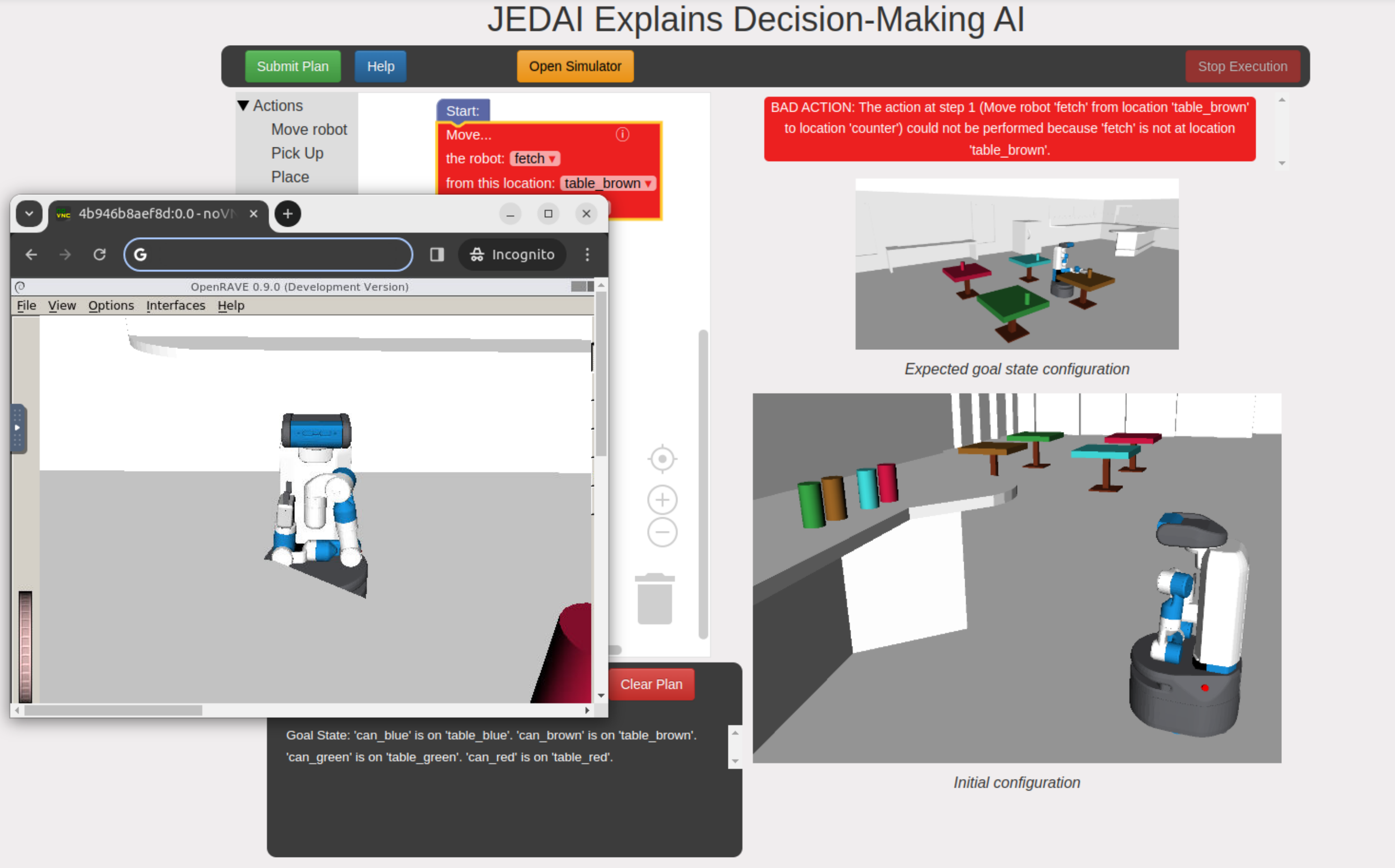}
            \caption{Screenshot of the JEDAI user interface.}
            \label{fig:unmod_jedai}
        \end{figure}

\end{landscape}


\end{document}